\documentclass[10pt,twocolumn,letterpaper]{article}

\usepackage{cvpr}              
\usepackage[accsupp]{axessibility}  

\usepackage{graphicx}
\usepackage{amsmath}
\usepackage{amssymb}
\usepackage{booktabs}
\usepackage{enumitem}
\usepackage{algorithm}
\usepackage{algcompatible}

\newtheorem{assumption}{Assumption}

\newtheorem{theorem}{Theorem}
\newtheorem{remark}{Remark}[theorem]

\usepackage[pagebackref,breaklinks,colorlinks]{hyperref}

\usepackage[capitalize]{cleveref}
\crefname{section}{Sec.}{Secs.}
\Crefname{section}{Section}{Sections}
\Crefname{table}{Table}{Tables}
\crefname{table}{Tab.}{Tabs.}

\def\cvprPaperID{6023} 
\def\confName{CVPR}
\def\confYear{2022}

\begin{document}

\title{Closing the Generalization Gap of Cross-silo Federated Medical Image Segmentation}

\author{
An Xu\thanks{Work done during an internship at NVIDIA. A.X. and H.H. were partially supported by NSF IIS 1845666, 1852606, 1838627, 1837956, 1956002, IIA 2040588. Implementation of this work is available at https://nvidia.github.io/NVFlare/research/fed-sm} \\ Univ. of Pittsburgh
\and
Wenqi Li \\ NVIDIA
\and
Pengfei Guo \\ Johns Hopkins University
\and
Dong Yang \\ NVIDIA
\and
Holger Roth \\ NVIDIA
\and
Ali Hatamizadeh \\ NVIDIA
\and
Can Zhao \\ NVIDIA
\and
Daguang Xu \\ NVIDIA
\and
Heng Huang \\ Univ. of Pittsburgh
\and
Ziyue Xu \\ NVIDIA}

\maketitle

\begin{abstract}
Cross-silo federated learning (FL) has attracted much attention in medical imaging analysis with deep learning in recent years as it can resolve the critical issues of insufficient data, data privacy, and training efficiency. However, there can be a generalization gap between the model trained from FL and the one from centralized training. This important issue comes from the non-iid data distribution of the local data in the participating clients and is well-known as client drift. In this work, we propose a novel training framework FedSM to avoid the client drift issue and successfully close the generalization gap compared with the centralized training for medical image segmentation tasks for the first time. We also propose a novel personalized FL objective formulation and a new method SoftPull to solve it in our proposed framework FedSM. We conduct rigorous theoretical analysis to guarantee its convergence for optimizing the non-convex smooth objective function. Real-world medical image segmentation experiments using deep FL validate the motivations and effectiveness of our proposed method.
\end{abstract}

\section{Introduction}

Deep learning models have shown success in computer vision tasks in recent years \cite{he2016deep,simonyan2014very,ronneberger2015u}. However, training deep models that generalize well on unseen test data may require massive training data. Unfortunately, we are usually faced with \textbf{insufficient data} in a single medical institution for the medical image segmentation task due to the expensive procedure of collecting enough patients' data with experts' labeling.

A straightforward solution to address the insufficient data issue is gathering data from all the available medical institutions, while the amount of data owned by any single institution may be insufficient to train a well-performing deep model. However, this approach will raise the concern for \textbf{data privacy}. On one hand, collecting medical data is expensive as mentioned above, and those data have become a valuable asset at a medical institution. Institutions with more data may be more reluctant to contribute their data. In addition, medical institutions bear the obligation to keep the data collected from patients secure. Gathering data may expose patients to the risk of data leakage.

Of course, we can leverage the existing vanilla distributed training method \cite{li2014scaling,xu2021step,xu2020acceleration} to keep the institution's data local and share only the gradient with a central server. But the training of deep model requires many iterations to converge, leading to unacceptable \textbf{communication complexity} for vanilla distributed training. It is not secure neither as recent works \cite{zhu2019deep,zhao2020idlg,geiping2020inverting,yin2021see} have shown that pixel-level images can be recovered from the leaked gradient.

Recently, federated learning (FL) \cite{konevcny2016federated,gao2021convergence,xu2021double,gu2021privacy} have been proposed to tackle all the above issues (insufficient data, data privacy, training efficiency). In medical applications, we are most interested in the cross-silo federated learning where we have a limited number of participating clients compared with cross-device federated learning (e.g., mobile devices) \cite{kairouz2019advances,liu2021feddg,guo2021multi}. Specifically, in each training round of FedAvg \cite{mcmahan2017communication}, the \textit{de facto} algorithm for FL, each client will perform local training with the global model received from a central server for multiple iterations. Then the server gathers all the local models from each client and averages them as the new global model. Nevertheless, for FedAvg and its variants, a non-negligible issue called ``client drift" arises due to non-iid data distribution on different clients. The local models on different clients will gradually diverge from each other during the local training. Client drift can drastically jeopardize the training performance of the global model when the data similarity decreases (more non-iid) \cite{hsieh2020non,hsu2019measuring}. Theoretically, it leads to a convergence rate more sensitive to the number of local training steps \cite{yu2019linear}.

Throughout this paper, we refer to centralized training as gathering data from clients and then training the model. Note that centralized training is impractical as it violates data privacy, but offers a performance upper bound for FL algorithms. Despite numerous efforts and previous works, there is still a \textbf{generalization gap} between FL and the centralized training. In this paper, unlike any previous works, we propose a novel training framework called \underline{Fed}erated \underline{S}uper \underline{M}odel (FedSM) to avoid confronting the difficult client drift issue at all for FL medical image segmentation tasks. In FedSM, instead of finding one global model that fits all clients' data distribution, we propose to produce personalized models to fit different data distributions well and a novel model selector to decide the closest model/data distribution for any test data.

We summarize our contributions as follows.
\begin{itemize}[leftmargin=0.18in]
\setlength{\itemsep}{0pt}
    \item We propose a novel training framework FedSM to avoid the client drift issue and close the generalization gap between FL and centralized training for medical segmentation tasks for the first time to the best of our knowledge.
    \item We propose a novel formulation for personalized FL optimization, and a novel personalized method called SoftPull to solve it in our framework FedSM. A rigorous convergence analysis with common assumptions in FL is given for the proposed method.
    \item Experiments in real-world FL medical image segmentation tasks validate our motivation and the superiority of our methods over existing FL baselines.
\end{itemize}

\section{Related Works}

Here we introduce existing different approaches to improve the model performance in FL with representative methods. First, the FL optimization problem is usually defined as $\min_{w} \frac{1}{K}\sum^{K}_{k=1} p_k L_{\mathcal{D}_k}(w)$, where the coefficient $p_k=\frac{n_k}{n}$, $n_k$ is the number of client $k$'s data, and the total number of data $n=\sum^{K}_{k=1}n_k$. $L_{\mathcal{D}_k}$ is the objective at client $k$ with its local data $\mathcal{D}_k$, and $w$ is the model weights.

\textbf{FedAvg.} In FedAvg, clients will receive the starting model $w_r$ from the server at training round $r$. Each client $k$ performs $E$ epochs of local training to update the local model to $w_{r+1}^{(k)}$ with the popular momentum SGD or Adam \cite{kingma2014adam} optimizer depending on the application needs. Then the server gathers and averages the local models to $w_{r+1}=\frac{1}{K}\sum^{K}_{k=1}p_k w^{(k)}_{r+1}$.

\textbf{Restrict Local Training.} To discourage the local models from diverging due to non-iid data distribution, FedProx \cite{li2018federated} proposes to add a proximal loss term $\|w^{(k)}_{r+1}-w_{r}\|^2_2$ to the objective function for client $k$. It implies that the local training will encourage $w^{(k)}_{r+1}$ to stay close to the starting point $w_r$, such that $\{w^{(k)}_{r+1}\}_{k\in\{1,2,\cdots,K\}}$ will be close to each other to alleviate the client drift issue.

\textbf{Correct Client Drift.} Motivated by variance reduction techniques in optimization such as SVRG \cite{johnson2013accelerating}, SAGA \cite{defazio2014saga}, inter-client variance reduction techniques \cite{acar2020federated,karimireddy2020scaffold,liang2019variance} are proposed for FL by correcting the local training with the predicted local and global updating direction. These methods are usually tested with convex or simple non-convex models/objectives. For the practical training of complicated deep models, \cite{defazio2018ineffectiveness} shows that variance reduction techniques fail to perform well in that correcting the stochastic gradient with variance reduction usually does not hold in deep learning due to common augmentation tricks such as batch normalization \cite{ioffe2015batch} and dropout \cite{srivastava2014dropout}, etc.

\textbf{Personalization.} Personalized models are usually a fine-tuned version of the global model to better fit the local data distribution of a specific client. We can fine-tune the global model \cite{wang2019federated} on a client's local data like the local training, or following MAML-based personalized methods \cite{t2020personalized,fallah2020personalized,jiang2019improving}. However, an intrinsic drawback of the personalized models is that they generalize poorly on other sites' data and unseen data. In this work, we focus on finding a model that generalizes as well as centralized training for all clients.

\textbf{Other Topics.} There are also many other emerging and interesting topics in FL, such as heterogeneous optimization \cite{wang2020tackling,li2018federated}, fairness and robustness \cite{mohri2019agnostic,Li2020Fair,li2021ditto}, clustered federated learning \cite{ghosh2020efficient}, etc. These topics are not directly related to our work but can be valuable for potential future extension. A recent work FedDG \cite{liu2021feddg} requires sharing partial information of the data, therefore it breaks the data privacy constraint to some extent. In this work, we share only the model update information for maximal data privacy.

\section{Methodology}
In this section, we present our motivation and the proposed method that can close the generalization gap for FL medical image segmentation tasks in detail.

\textbf{Motivation.} In traditional FL, the goal is to collaboratively train one global model that generalizes well on all clients' joint data distribution. The client drift issue comes from the fact that we only have access to clients' local data distribution during the local training. It is hard to train a global model generalizing as well as centralized training due to this issue despite numerous existing works. In this work, however, we show that it is possible to get rid of the client drift issue. Specifically, we propose that
\begin{itemize}[leftmargin=0.18in]
\setlength{\itemsep}{0pt}
    \item for the test data, we search for the closest (i.e., the most similar) local data distribution from all clients (Section \ref{sec:fedsm}).
    \item we find a model with the best generalization performance on this selected local data distribution, and use it for the inference of the test data (Section \ref{sec:softpull}).
\end{itemize}

\subsection{New Framework: FedSM}\label{sec:fedsm}

\begin{figure}
    \centering
    \includegraphics[width=.8\columnwidth]{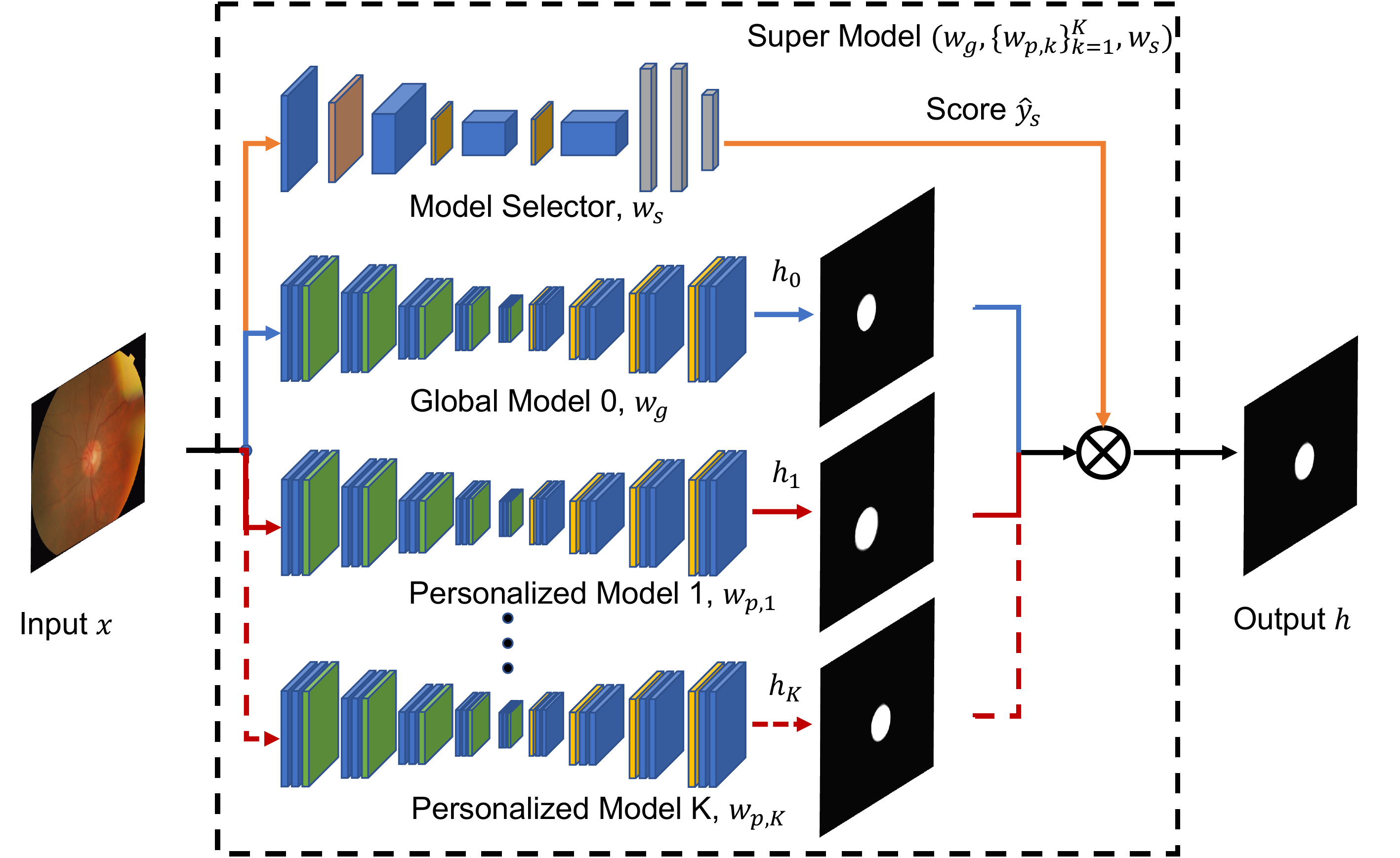}
    \caption{The proposed FedSM framework with ``super model".}
    \label{fig:fedsm framework}
\end{figure}

The first motivation above motivates us to design a new and general FL framework FedSM, where we train a \underline{Fed}erated ``\underline{S}uper \underline{M}odel" consisting of the global model, personalized models, and a model selector. These components are illustrated in Figure \ref{fig:fedsm framework} and we elaborate them as follows.

Global model $w_g$: the global model trained by FedAvg. It generalizes better than personalized models on the joint data distribution of all clients, but there is still a gap compared with centralized training. Suppose the model function is $f$ and we denote its output as $h_0=f(w_g,x)$ for data $x$.

Personalized models $w_{p,k}$: the personalized models trained by any personalization FL training method. A personalized model usually generalizes better on local data than the global model. We denote its output as $h_k=f(w_{p,k},x)$, where $k\in \{1,2,\cdots,K\}$.

Model selector $w_s$: its goal is to determine the match between the unseen data input $x$ and each of the global/personalized models for inference. Specifically, it outputs a normalized prediction score vector $\widehat{y}_s$. The final output $h$ is determined by $\widehat{y}_s$ and $[h_0,h_1,\cdots,h_K]$. Suppose the candidate model set $\Omega\subseteq \{0,1,2,\cdots,K\}$, then $\sum_{k\in\Omega}\widehat{y}_{s,k}=1$ and $h=\sum_{k\in\Omega}\widehat{y}_{s,k}h_k$. We discuss the potential training methods as follows.

\subsubsection{Ensemble}

Suppose we already have the trained global model and personalized models. Given the FedSM framework as shown in Figure \ref{fig:fedsm framework}, a straightforward approach is to ensemble the outputs $[h_0,h_1,\cdots,h_K]$ from all models as the final output $h=\sum^{K}_{k=0}\widehat{y}_{s,k}h_k$. Let the ground truth of data $x$ be $y$ and the loss function be $L$. Then, we compute the loss $L(h,y)$ and update the model selector $w_s$ via FedAvg.

However, in practice we find it \textbf{hard to train} the model selector in this way in FL. The final performance can be even inferior to the global model. Let the desired value $y_s=\min_{\widehat{y}_s}L(\sum^{K}_{k=0}\widehat{y}_{s,k}h_k,y)$. We found that it was caused by the difficulty to train $\widehat{y}_s$ to the desired value $y_s$ by $\min_{w_s}L(\sum^{K}_{k=0}\widehat{y}_{s,k}h_k,y)$ as $w_s$ is the model weights to optimize. For each data input $x$, we may need many training steps to $\min_{w_s}L(\sum^{K}_{k=0}\widehat{y}_{s,k}h_k,y)$ such that $\widehat{y}_s$ will be close to $y_s$. However, it is unacceptable due to the large amount of computation cost.

Another issue of this approach is that we cannot start training the model selector until the training of the global model and personalized models finishes, which incurs \textbf{extra communication rounds} for FL.

\subsubsection{FedSM-extra}

To tackle the \textbf{training difficulty} in ensemble, here we propose to compute
\begin{equation}\label{eq:fedsm-extra}
    y_s=one\_hot(\arg \min_k\{L(h,h_k)\}_{k=0}^{K})\,,
\end{equation}
where ``one\_hot'' denotes one hot encoding. Then we compute the cross entropy loss $L_s(\widehat{y}_s,y_s)$ to update the model selector. In this way, the model selector is more clear about the desired value $y_s$. Thus it will be easier to train. We refer to this approach as FedSM-extra as it still needs extra communication rounds like the ensemble approach.

\subsubsection{FedSM}

To address the issue of \textbf{extra training rounds}, the model selector needs to be trained together with the global model and personalized models. Nevertheless, from Eq.~(\ref{eq:fedsm-extra}) we can see that the desired $y_s$ depends on the output of the trained global model and personalized models. Therefore, we need to decouple their dependency. As a further simplification, suppose the training data $x$ comes from  the client $k\in\{1,2,\cdots,K\}$, here we propose
\begin{equation}\label{eq:fedsm}
    y_s=one\_hot(k)\,.
\end{equation}
Intuitively, the personalized model $k$ tends to generalize better on client $k$'s own local data. It is safe to set $y_s$ as the corresponding client index. Though theoretically, it may degrade the performance of Eq.~(\ref{eq:fedsm-extra}), it is more practical due to no extra training rounds. We refer to this approach as FedSM which addresses all the issues raised by the ensemble.

\subsection{New Personalization: SoftPull} \label{sec:softpull}

In this section, we present a new personalized FL optimization formulation and a method, SoftPull, to solve it and produce personalized models for FedSM. We first present existing interpolation methods to tackle the insufficient local data issue.

Let the global dataset be $\mathcal{D}$. To tackle the insufficient local data issue, \cite{mansour2020three} proposes dataset interpolation for each client as $\min_{w_{p,k}} \lambda L_{\mathcal{D}_k}(w_{p,k}) + (1-\lambda)L_{\mathcal{D}}(w_{p,k})$, where coefficient $\lambda\in[0,1]$. As client $k\in\{1,2,\cdots,K\}$, it leads to $K$ optimization problems and is inefficient to solve. Besides, it is hard to acquire the information of the global dataset $D$ during the local training. \cite{mansour2020three} also proposes model interpolation $\min_{w_g,w_{p,k},\lambda} \sum^{K}_{k=1}L_{\mathcal{D}_k}(\lambda w_{p,k}+(1-\lambda)w_{g})$. To efficiently solve the model interpolation problem, APFL \cite{deng2020adaptive} proposes
\begin{align}
    w_g^* &= \arg\min_{w_g} L_{\mathcal{D}}(w_g)\,,\\
    w_{p,k}^* &= \arg\min L_{\mathcal{D}_k}(\lambda w_{p,k} + (1-\lambda)w_g^*)\,,\\
    w_{p.k} &\leftarrow \lambda w_{p,k}^* + (1-\lambda)w_g^*\,.
\end{align}

\textbf{Motivation.} We observe that model interpolation tries to find an appropriate combination between the FL global and local models. When the local data distribution is not similar to the global data distribution at all, we expect $\lambda\rightarrow 1$. When they are similar, we expect $\lambda \rightarrow \frac{1}{K}$ to leverage the global data information to improve the local generalization as the local dataset is small. Nevertheless, the formulation of APFL has two potential drawbacks:
\begin{itemize}[leftmargin=0.18in]
\setlength{\itemsep}{0pt}
    \item The involved global model $w^*_g$ may not generalize well on $\mathcal{D}$ and $\mathcal{D}_k$, but will affect the FL training.
    \item What objective function it is exactly optimizing is not clear.
\end{itemize}
In our problem formulation, we first suppose $w^*_k$ is the local optimum of client $k$:
\begin{equation}\label{eq:local optimum}
    w_k^* = \arg\min_w L_{\mathcal{D}_k}(w)\,.
\end{equation}
However, local optimum $w^*_k$ may not generalizes well due to lack of local training data. Instead of interpolating the global and local optimum, we propose that the desired \textit{personalized optimum $w^*_{p,k}$ is an interpolation between the local optimum of client $k$ and other clients' personalized optima}:
\begin{equation}\label{eq:new interpolation}
    w^*_{p,k}=\lambda w^*_k + (1-\lambda)\frac{1}{K-1}\sum^{K}_{k^\prime=1,k^\prime\neq k}w^*_{p,k^\prime}\,.
\end{equation}
The new interpolation avoids the global model and guarantees that the interpolated model is the optimum to some explicit objective function, as opposed to APFL. In fact, the personalized optimum $w^*_{p,k}$ is also an interpolation between the local optimum of client $k$ and other clients' local optimum because Eq.~(\ref{eq:new interpolation}) is identical to
\begin{equation}
    w^*_{p,k}=\lambda w^*_k + (1-\lambda)\frac{1}{K-1}\sum^{K}_{k^\prime=1,k^\prime\neq k}w^*_{k^\prime}\,.
\end{equation}
However, Eq.~(\ref{eq:new interpolation}) is better to help us to find what objective function we are optimizing as we can turn it to
\begin{equation}
    w^*_k=\frac{1}{\lambda}w^*_{p,k} - \frac{1-\lambda}{\lambda}\frac{1}{K-1}\sum^{K}_{k^\prime=1,k^\prime\neq k}w^*_{p,k^\prime}\,.
\end{equation}
Compare it with Eq.~(\ref{eq:local optimum}) and we immediately have $\{w^*_{p,k}\}^{K}_{k=1}$ as the solution to the optimization problem
\begin{equation}\label{eq:new opt prob}
    \min_{\{w_{p,k}\}}\sum^{K}_{k=1}L_{\mathcal{D}_k}(\frac{1}{\lambda}w_{p,k}-\frac{1-\lambda}{\lambda}\frac{1}{K-1}\sum^{K}_{k^\prime=1,k^\prime\neq k}w_{p,k^\prime})\,.
\end{equation}
To solve the proposed new personalized FL optimization problem Eq.~(\ref{eq:new opt prob}), we propose a new method, SoftPull ($\lambda \in [\frac{1}{K},1]$), with the simplification of substituting $w^*_k$ with the locally trained model in Eq.~(\ref{eq:new interpolation}), that is, after each training round at the server,
\begin{equation}\label{eq:softpull}
    w_{p,k} \leftarrow \lambda w_{p,k} + (1-\lambda)\frac{1}{K-1}\sum^{K}_{k^\prime=1,k^\prime\neq k} w_{p,k^\prime}\,.
\end{equation}
The corresponding algorithm is summarized in Algorithm \ref{alg:fedsm training}, line 16. When $\lambda=\frac{1}{K}$, it reduces to the ``hard'' averaging in FedAvg. To analyze the convergence, we start with common assumptions as follows.
\begin{assumption}\label{lipschitz}
(Lipschitz Smooth) The loss function $L_{\mathcal{D}_k}$ is $L$-smooth, that is, $\forall w_1, w_2 \in \mathbb{R}^d$, we have
\begin{equation}
    \|\nabla L_{\mathcal{D}_k}(w_1) - \nabla L_{\mathcal{D}_k}(w_2)\|^2_2 \leq L\|w_1 - w_2\|^2_2\,.
\end{equation}
\end{assumption}
\begin{assumption}\label{bounded variance}
(Bounded Variance) The stochastic gradient $\nabla L_{\mathcal{D}_k}(w,x)$ has bounded variance $\forall w\in\mathbb{R}^d$:
\begin{equation}
    \mathbb{E}\|\nabla L_{\mathcal{D}_k}(w,x)-\nabla L_{\mathcal{D}_k}(w)\|^2_2\leq \sigma^2\,.
\end{equation}
where $\mathbb{E}$ is an expectation over $x\in \mathcal{D}_k$.
\end{assumption}
\begin{assumption}\label{bounded gradient}
\cite{reddi2020adaptive} The gradient $\nabla L_{\mathcal{D}_k}(w)$ has bounded value $\forall w\in\mathbb{R}^d$: $\|\nabla L_{\mathcal{D}_k}(w)\|^2_2\leq G^2$.
\end{assumption}
\begin{theorem}\label{convergence}
Suppose Assumptions \ref{lipschitz}, \ref{bounded variance}, and \ref{bounded gradient} exist. Let the proposed objective in Eq.~(\ref{eq:new opt prob}) be $F$, superscript $(r,m)$ denote the global iteration, and $\overline{w}$ denote the average, then
\begin{align}
    &\frac{1}{KRM}\sum^{R-1}_{r=0}\sum^{M-1}_{m=0}\sum^{K}_{k=1}\mathbb{E}\|\nabla_{w_{p,k}^{r,m}}F\|^2_2\\
    &= \mathcal{O}(\frac{1}{\eta RM\lambda^2} + \frac{(1-\lambda)^2}{KRM\eta^2\lambda^2}\sum^{K}_{K=1}\sum^{R-1}_{r=0}\mathbb{E}\|w^{r,M}_{p,k}-\overline{w}^{r,M}_{p,k}\|^2_2 \notag\\
    &\quad + \frac{(1-\lambda)^2}{KRM\lambda^4}\sum^{K}_{k=1}\sum^{R-1}_{r=0}\sum^{M-1}_{m=0}\mathbb{E}\|w^{r,m}_{p,k}-\overline{w}^{r,m}_{p,k}\|^2_2) \notag\\
    &=\mathcal{O}(\frac{1}{\eta RM\lambda^2} + \frac{M\sum^{R-1}_{r=0}(1-\lambda)^2}{R\lambda^2} + \frac{M^2\eta^2\sum^{R-1}_{r=0}(1-\lambda)^2}{R\lambda^4})\,. \notag
\end{align}
If $\eta=\mathcal{O}(\frac{1}{\sqrt{RM}})$ and $M=\mathcal{O}(R^{\frac{1}{3}})$, its convergence rate is $\mathcal{O}(\frac{1}{\sqrt{RM}})$ with a convergence error $\mathcal{O}(\frac{M\sum^{R-1}_{r=0}(1-\lambda)^2}{R\lambda^2})$.
\end{theorem}
\begin{remark}\label{remark:sim}
When the data similarity is low among clients, we should set a larger $\lambda$ to reduce the effect of $\|w^{r,m}_{p,k}-\overline{w}^{r,m}_{p,k}\|^2_2$ and ensure the convergence rate. It is intuitively valid as the client has less to learn from other clients.
\end{remark}
\begin{remark}\label{remark:error}
$\lambda \downarrow$ and the convergence error $\uparrow$, but it does not mean worse generalization because we do not want to overfit local data. We will empirically tune and validate it.
\end{remark}
The proof can be found in Appendix \ref{appendix:convergence}.

\subsection{All Together}

\begin{algorithm}[t]
\caption{FedSM training.}\label{alg:fedsm training}
\begin{algorithmic}[1]
    \STATE \textbf{Input:} local dataset $\mathcal{D}_k$, rounds $R$, number of sites $K$, learning rate $\eta$, $\eta_s$, coefficient $\lambda$, client weight $\frac{n_k}{n}$.
    \STATE \textbf{Initialize:} global model $w_g$, personalized model $w_{p,k}$, model selector $w_s$, base optimizer $\text{OPT}(\cdot)$
    
    \FOR{round $r=1,2,\cdots,R$}
        \STATE SERVER: send models ($w_g$, $w_{p,k}$, $w_s$) to client $k$.
        \FOR{CLIENT $k\in\{1,2,\cdots,K\}$ in parallel}
            \STATE initialize $w_{g,k}\leftarrow w_g$, $w_{s,k} \leftarrow w_s$
            \FOR{batch $(x,y) \in \mathcal{D}_k$}
                \STATE $w_{g,k}\leftarrow \text{OPT}(w_{g,k},\eta,\nabla_{w_{g,k}} L(f(w_{g,k};x),y))$
                \STATE $w_{p,k}\leftarrow \text{OPT}(w_{p,k},\eta,\nabla_{w_{p,k}} L(f(w_{p,k};x),y))$
                \STATE // \textit{$y_s$ from Eq.~(\ref{eq:fedsm})}
                \STATE $w_{s,k}\leftarrow \text{OPT}(w_{s,k}, \eta_s,\nabla_{w_{s,k}}L_s(f_s(w_{s,k};x),y_s))$
            \ENDFOR
            \STATE send ($w_{g,k}$, $w_{p,k}$, $w_{s,k}$) to server
        \ENDFOR
        \STATE SERVER: $w_g, w_s\leftarrow \sum^{K}_{k=1}\frac{n_k}{n} w_{g,k}, \sum^{K}_{k=1}\frac{n_k}{n} w_{s,k}$
        \STATE SERVER: $\forall k \in \{1,2,\cdots,K\}$, $w_{p,k}\leftarrow \lambda w_{p,k} + (1-\lambda)\frac{1}{K-1}\sum^{K}_{k^\prime=1, k^\prime\neq k} w_{p,k^\prime}$ \hfill // \textit{SoftPull}
    \ENDFOR
    
    \STATE \textbf{Output:} model ($w_g$, $\{w_{p,k}\}^{K}_{k=1}$, $w_s$)
\end{algorithmic}
\end{algorithm}

\begin{algorithm}[t]
\caption{FedSM inference.}\label{alg:fedsm inference}
\begin{algorithmic}[1]
    \STATE \textbf{Input:} data $x$, model ($w_g$, $\{w_{p,k}\}^{K}_{k=1}$, $w_s$), threshold $\gamma$
    
    \STATE $\widehat{y}_s = f_s(w_s; x)$
    \IF {$\max (\widehat{y}_s)>\gamma$}
        \STATE $k=\arg\max(\widehat{y}_s)\in\{1,2,\cdots,K\}$// \textit{high confidence}
        \STATE $\widehat{y}=f(w_{p,k};x)$
    \ELSE
        \STATE $\widehat{y}=f(w_g;x)$ \hfill // \textit{low confidence}
    \ENDIF
    
    \STATE \textbf{Output:} $\widehat{y}$
\end{algorithmic}
\end{algorithm}

We summarize the proposed SoftPull method to train personalized models and the FedSM framework consisting of the model selector, global model, and personalized models in Algorithm \ref{alg:fedsm training}. Compared with FedAvg, the communication cost of each training round is $2w_g+w_s$ for FedSM. We note that some methods such as Scaffold \cite{karimireddy2020scaffold} have a cost of $2w_g$. After the training, the server sends the super model ($w_g$, $\{w_{p,k}\}^{K}_{k=1}$, $w_s$) to each client for inference, which incurs only a one-time communication cost.

For the FedSM inference in Algorithm \ref{alg:fedsm inference}, we propose a heuristic technique that the model selector selects the global model when its confidence is low, because we do not have label 0 in Eq.~(\ref{eq:fedsm}) (the global model) during training. Intuitively, if the test data is not similar to any local data distribution, the global model should be a better choice for its inference, in that it covers the joint data distribution while the personalized model covers only one local data distribution. It also guarantees that FedSM is at least not worse than the global model from FedAvg with an appropriate threshold $\gamma$.

For FedSM-extra, both the training and inference algorithms are the same except for the determination of $y_s$, the extra training rounds, and no need for the threshold $\gamma$. More details are available in Appendix \ref{appendix: fedsm-extra}.

\section{Experiments}

\begin{table}[t]
\small
    \centering
    \begin{tabular}{c|cccccc|c}
        \toprule
        Client & 1 & 2 & 3 & 4 & 5 & 6 & Global\\
        \midrule
        Train & 50 & 98 & 47 & 230 & 80 & 400 & 905\\
        Val & 25 & 49 & 24 & 115 & 40 & 200 & 453 \\
        Test & 26 & 48 & 23 & 115 & 39 & 200 & 451 \\
        \bottomrule
    \end{tabular}
    \caption{Retinal Dataset: number of data (2D image) in each client. The data sources from client 1 to 6 are Drishti-GS1 \cite{6867807}, RIGA \cite{almazroa2018retinal} BinRushed, RIGA Magrabia, RIGA MESSIDOR, RIM-ONE \cite{fumero2011rim}, and REFUGE \cite{refuge} respectively. Global refers to the data from all clients.}
    \label{tab:retinal num}
\end{table}

\begin{table}[t]
\small
    \centering
    \begin{tabular}{c|cccccc|c}
        \toprule
        Client & 1 & 2 & 3 & 4 & 5 & 6 & Global \\
        \midrule
        Train & 153 & 404 & 464 & 361 & 609 & 1179 & 3170 \\
        Val & 77 & 215 & 219 & 162 & 289 & 582 & 1544 \\
        Test & 61 & 245 & 198 & 150 & 329 & 532 & 1515 \\
        \bottomrule
    \end{tabular}
    \caption{Prostate Dataset: number of data (2D slices) in each client. The data sources from client 1 to 6 are I2CVB \cite{lemaitre2015computer}, MSD \cite{antonelli2021medical}, NCI\_ISBI\_3T, NCI\_ISBI\_DX \cite{nci-isbi}, Promise12 \cite{promise12}, and ProstateX \cite{prostatex} respectively.  Global refers to the data from all clients.}
    \label{tab:prostate num 2d}
\end{table}

\begin{figure*}[t]
    \centering
    \includegraphics[width=0.3\textwidth]{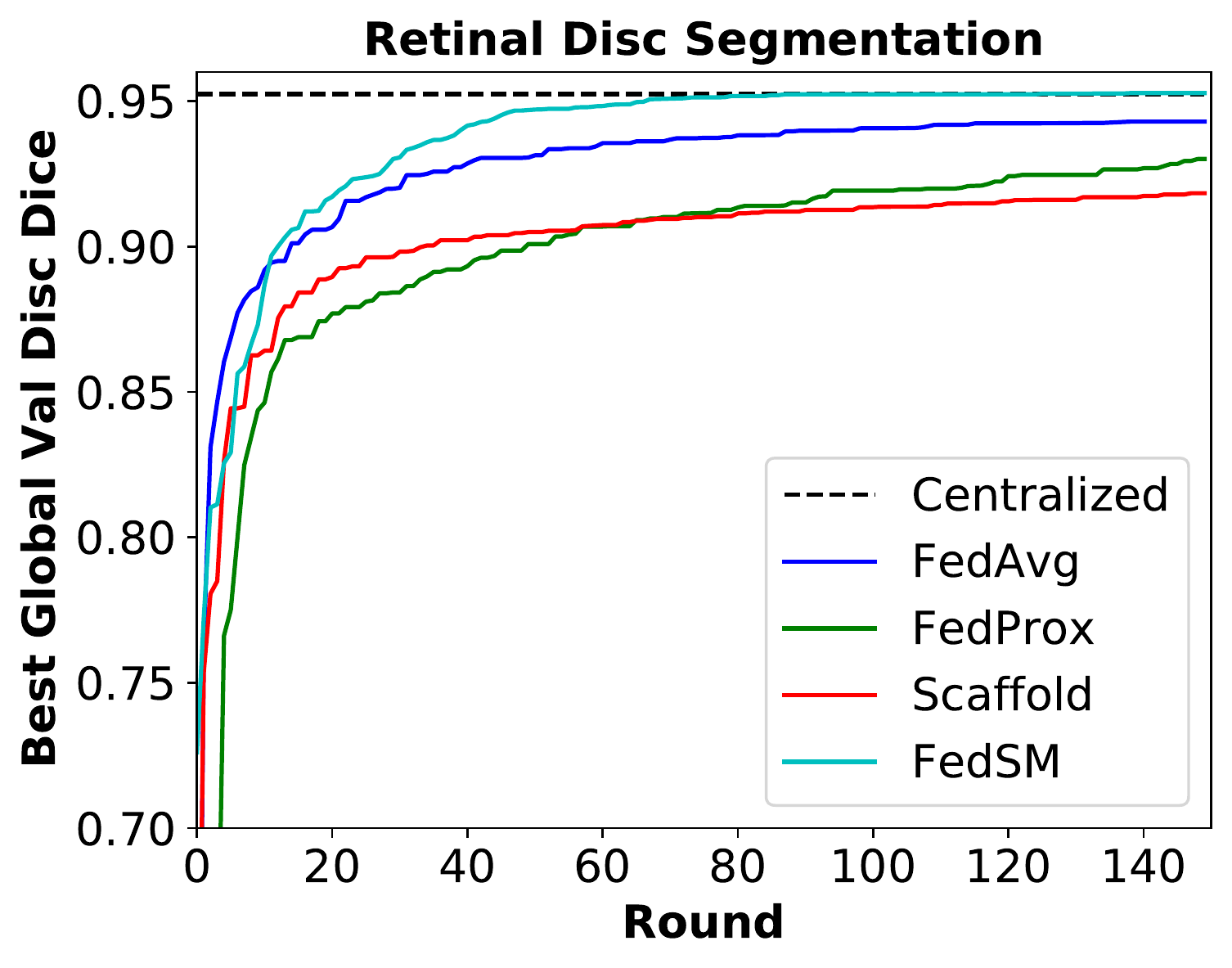}
    \includegraphics[width=0.3\textwidth]{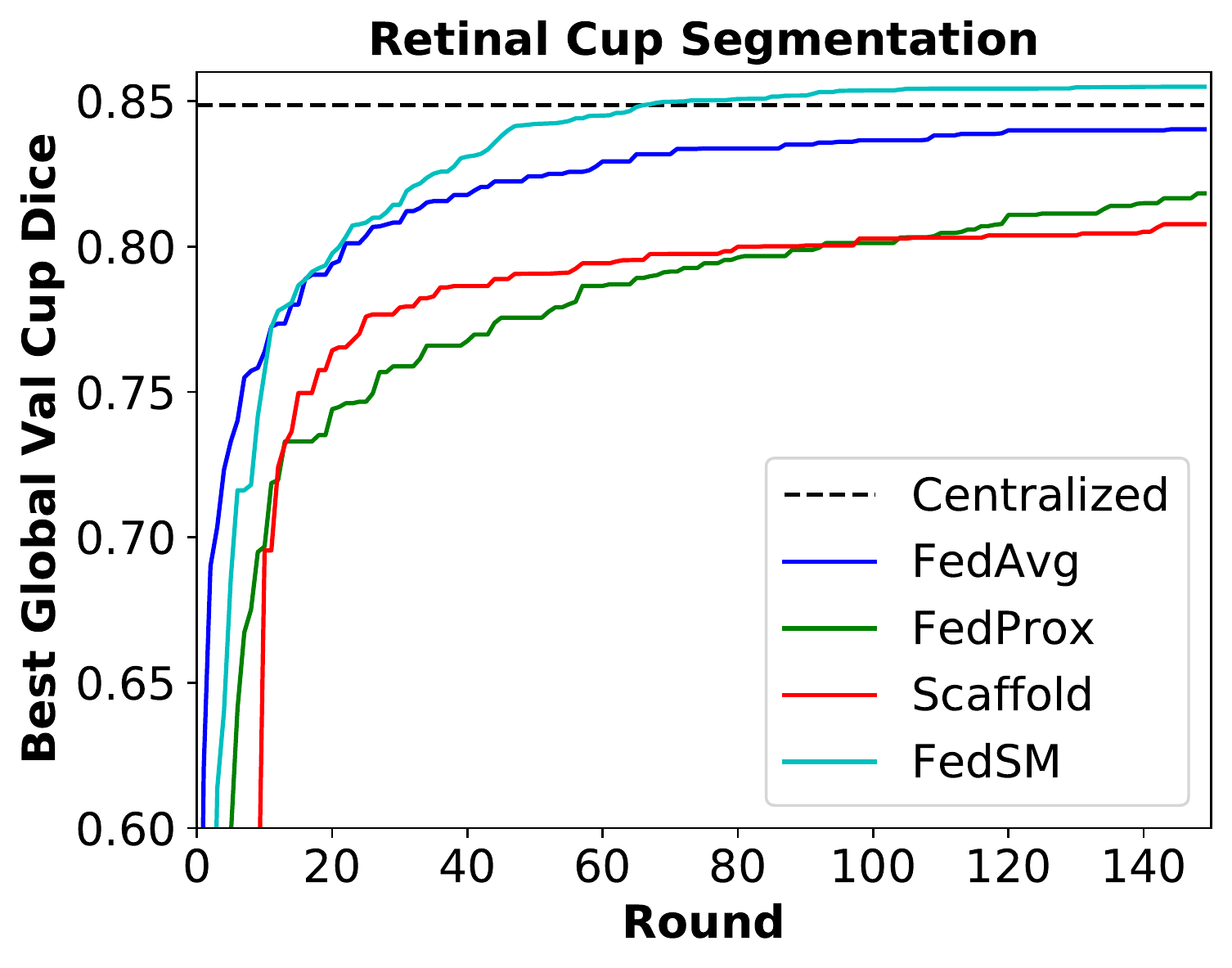}
    \includegraphics[width=0.3\textwidth]{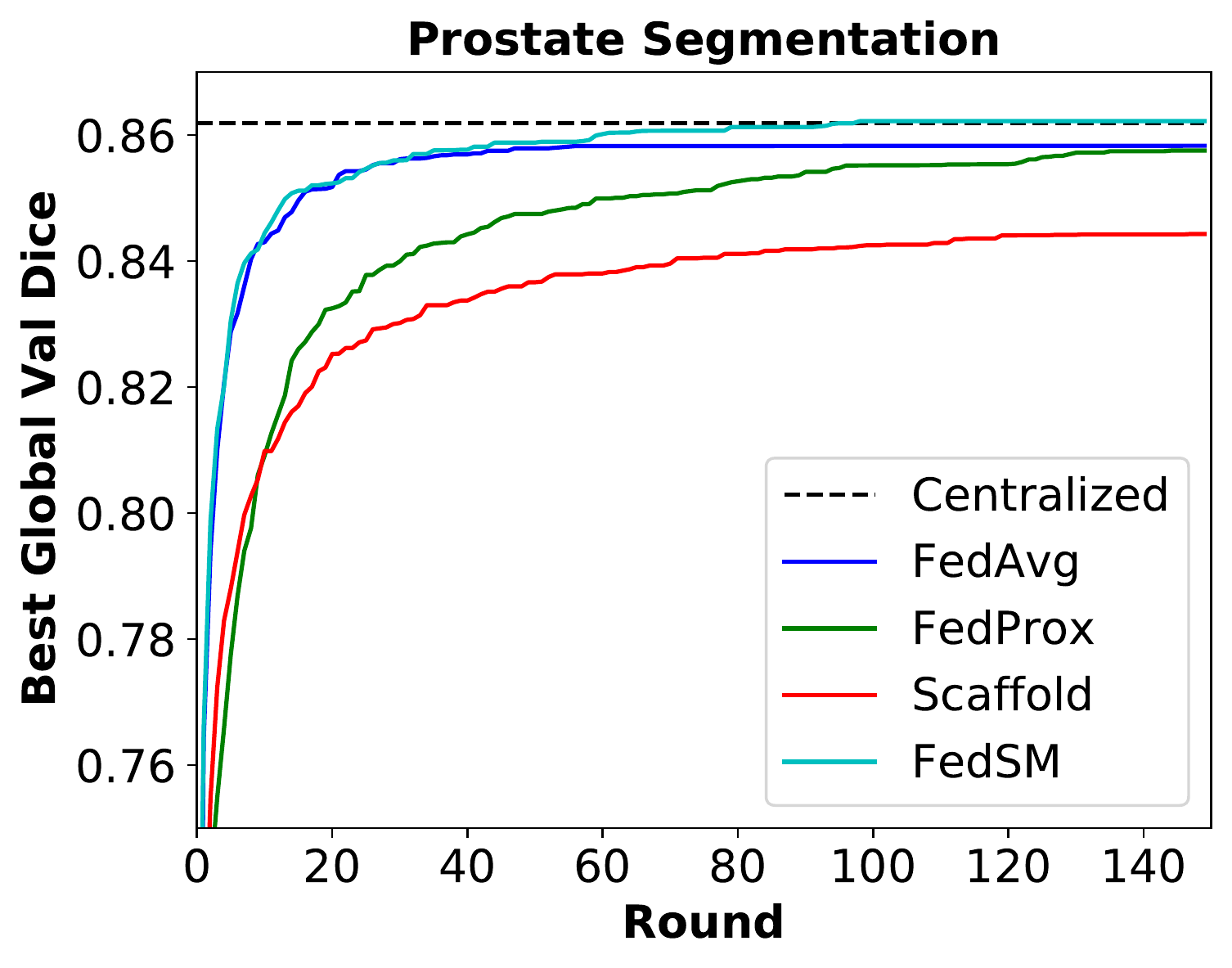}
    \caption{Training curves comparison. The curves are non-decreasing because we record the best result during training.}
    \label{fig:curve}
\end{figure*}

\begin{table*}[t]
\small
    \centering
    \begin{tabular}{c|cccccc|cc}
        \toprule
        Method & Client 1 & Client 2 & Client 3 & Client 4 & Client 5 & Client 6 & Client Avg Dice & Global Dice \\
        \midrule
        Centralized & 0.9161 & 0.8760 & 0.8758 & 0.9022 & 0.8510 & 0.9179 & 0.8898 & 0.9014 \\
        \midrule
        Client 1 Local & 0.8835 & 0.3331 & 0.7345 & 0.4933 & 0.3408 & 0.7015 & 0.5811 & 0.5902 \\
        Client 2 Local & 0.2346 & 0.8620 & 0.0886 & 0.7751 & 0.1791 & 0.4106 & 0.4250 & 0.5050 \\
        Client 3 Local & 0.8337 & 0.3402 & 0.8766 & 0.6010 & 0.3644 & 0.7794 & 0.6326 & 0.6594 \\
        Client 4 Local & 0.5108 & 0.8574 & 0.3457 & 0.9008 & 0.2361 & 0.6822 & 0.5888 & 0.6910 \\
        Client 5 Local & 0.5241 & 0.1584 & 0.3953 & 0.2039 & 0.8223 & 0.6222 & 0.4544 & 0.4662 \\
        Client 6 Local & 0.7908 & 0.6649 & 0.7325 & 0.7681 & 0.3742 & 0.9150 & 0.7076 & 0.7877 \\
        \midrule
        FedAvg & 0.8847 & 0.8679 & 0.8667 & 0.9015 & 0.7877 & 0.9172 & 0.8710 & 0.8923 \\
        FedProx & 0.8635 & 0.8522 & 0.8547 & 0.8952 & 0.6852 & 0.9095 & 0.8434 & 0.8749 \\
        Scaffold & 0.8380 & 0.8513 & 0.8215 & 0.8935 & 0.5671 & 0.9130 & 0.8141 & 0.8625 \\
        \midrule
        FedSM & \textbf{0.9132} & \textbf{0.8769} & \textbf{0.8865} & \textbf{0.9041} & \textbf{0.8483} & \textbf{0.9195} & \textbf{0.8914} & \textbf{0.9028} \\
        FedSM-extra & \textbf{0.9134} & \textbf{0.8763} & \textbf{0.8841} & \textbf{0.9038} & \textbf{0.8483} & \textbf{0.9172} & \textbf{0.8905} & \textbf{0.9007} \\
        \bottomrule
    \end{tabular}
    \caption{(low data similarity) Test Dice coefficient comparison of retinal segmentation. ``Client $k$ Local" refers to local training on client $k$. The first row refers to the performance on client 1$\sim$6's test data, their average, and the performance on all clients' test data. We report the average of disc and cup Dice coefficients here. We bold the best FL numbers. See Appendix \ref{appendix:additional exp} for their separate numbers and the visual comparison of segmentation.}
    \label{tab:retinal avg dice}
\end{table*}

\begin{table*}[t]
\small
    \centering
    \begin{tabular}{c|cccccc|cc}
        \toprule
        Method & Client 1 & Client 2 & Client 3 & Client 4 & Client 5 & Client 6 & Client Avg Dice & Global Dice \\
        \midrule
        Centralized & 0.9018 & 0.8583 & 0.8702 & 0.8844 & 0.8800 & 0.8474 & 0.8737 & 0.8651 \\
        \midrule
        Client 1 Local & 0.8582 & 0.3886 & 0.4476 & 0.2849 & 0.3830 & 0.4697 & 0.4720 & 0.4336 \\
        Client 2 Local & 0.7166 & 0.7669 & 0.8317 & 0.7341 & 0.6156 & 0.7754 & 0.7401 & 0.7403 \\
        Client 3 Local & 0.6470 & 0.8541 & 0.8549 & 0.6735 & 0.6591 & 0.7519 & 0.7401 & 0.7496 \\
        Client 4 Local & 0.4515 & 0.6566 & 0.6700 & 0.8518 & 0.4558 & 0.6267 & 0.6187 & 0.6148 \\
        Client 5 Local & 0.8198 & 0.7751 & 0.8469 & 0.8029 & 0.8038 & 0.7928 & 0.8069 & 0.8016 \\
        Client 6 Local & 0.8555 & 0.7965 & 0.8260 & 0.7206 & 0.6478 & 0.8466 & 0.7822 & 0.7809 \\
        \midrule
        FedAvg & 0.8775 & 0.8575 & 0.8700 & 0.8802 & 0.8717 & 0.8532 & 0.8684 & 0.8638 \\
        FedProx & \textbf{0.8948} & 0.8511 & 0.8722 & 0.8803 & 0.8668 & 0.8513 & 0.8694 & 0.8621 \\
        Scaffold & 0.8500 & 0.8440 & 0.8570 & 0.8423 & 0.8431 & 0.8412 & 0.8463 & 0.8446 \\
        \midrule
        FedSM & \textbf{0.8946} & \textbf{0.8596} & \textbf{0.8786} & \textbf{0.8898} & \textbf{0.8817} & \textbf{0.8535} & \textbf{0.8763} & \textbf{0.8692} \\
        FedSM-extra & 0.8886 & \textbf{0.8584} & \textbf{0.8766} & \textbf{0.8880} & \textbf{0.8760} & \textbf{0.8542} & \textbf{0.8736} & \textbf{0.8673} \\
        \bottomrule
    \end{tabular}
    \caption{(high data similarity) Test Dice coefficient comparison of prostate segmentation. We bold the best FL numbers. See Appendix \ref{appendix:additional exp} for the visual comparison.}
    \label{tab:prostate dice}
\end{table*}

\begin{table*}[t]
\small
    \centering
    \begin{tabular}{c|c|ccccccc|c|r}
        \toprule
        Unseen Client $k$ & Threshold $\gamma$ & GM & PM1 & PM2 & PM3 & PM4 & PM5 & PM6 & Dice & Best $\gamma$, Dice \\
        \midrule
        Client $k=6$ & 0 & 0 & 0.02 & 0 & 0.35 & 0 & 0.63 & N/A & 0.8587 & 1, 0.8906 \\
        Client $k=5$ & 0 & 0 & 0.31 & 0.03 & 0 & 0.61 & N/A & 0.05 & 0.4015 & 0.9, 0.4304 \\
        Client $k=4$ & 0 & 0 & 0 & 1.00 & 0 & N/A & 0 & 0 & 0.8869 & $<$0.95, 0.8870 \\
        Client $k=3$ & 0 & 0 & 0 & 0.57 & N/A & 0 & 0 & 0.43 & 0.8441 & $<$0.9, 0.8446\\
        Client $k=2$ & 0 & 0 & 0 & N/A & 0 & 0.92 & 0.08 & 0 & 0.8409 & $<$1, 0.8409 \\
        Client $k=1$ & 0 & 0 & N/A & 0 & 1.00 & 0 & 0 & 0 & 0.8839 & $<$0.99, 0.8839 \\
        \bottomrule
    \end{tabular}
    \caption{(retinal segmentation, Dice = average of disc and cup Dice coefficients) Model selection frequency from the model selector when FL train with clients $\{1,2,\cdots,6\}/\{k\}$ and test on the \textbf{unseen} client $k\in\{1,2,\cdots,6\}$. From left to right, GM denotes the global model and PM denotes the personalized model $\{1,2,\cdots,6\}/\{k\}$. The model selection frequency with the best $\gamma$, and the more detailed Dice results can be found in Appendix \ref{appendix:additional exp}. Note GM is never selected as the Threshold $\gamma$ is intentionally set to 0.}
    \label{tab:model selector frequency}
    \vspace{-10pt}
\end{table*}

\begin{figure}[t]
    \centering
    \includegraphics[width=.6\columnwidth]{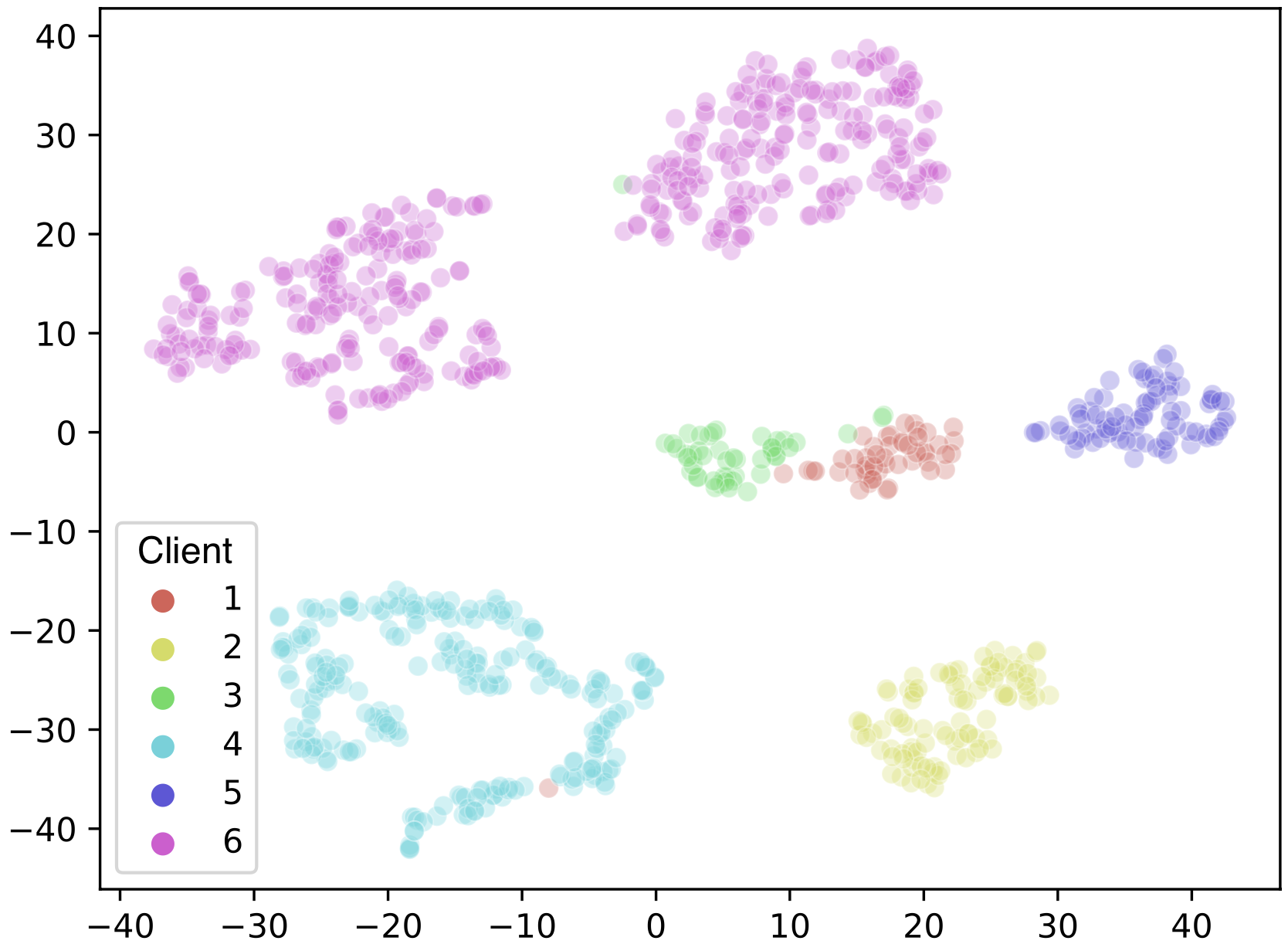}
    \caption{TSNE map of the features extracted form the model selector on retinal segmentation task.}
    \label{fig:tsne}
\end{figure}

\begin{figure*}
    \centering
    \includegraphics[width=0.3\textwidth]{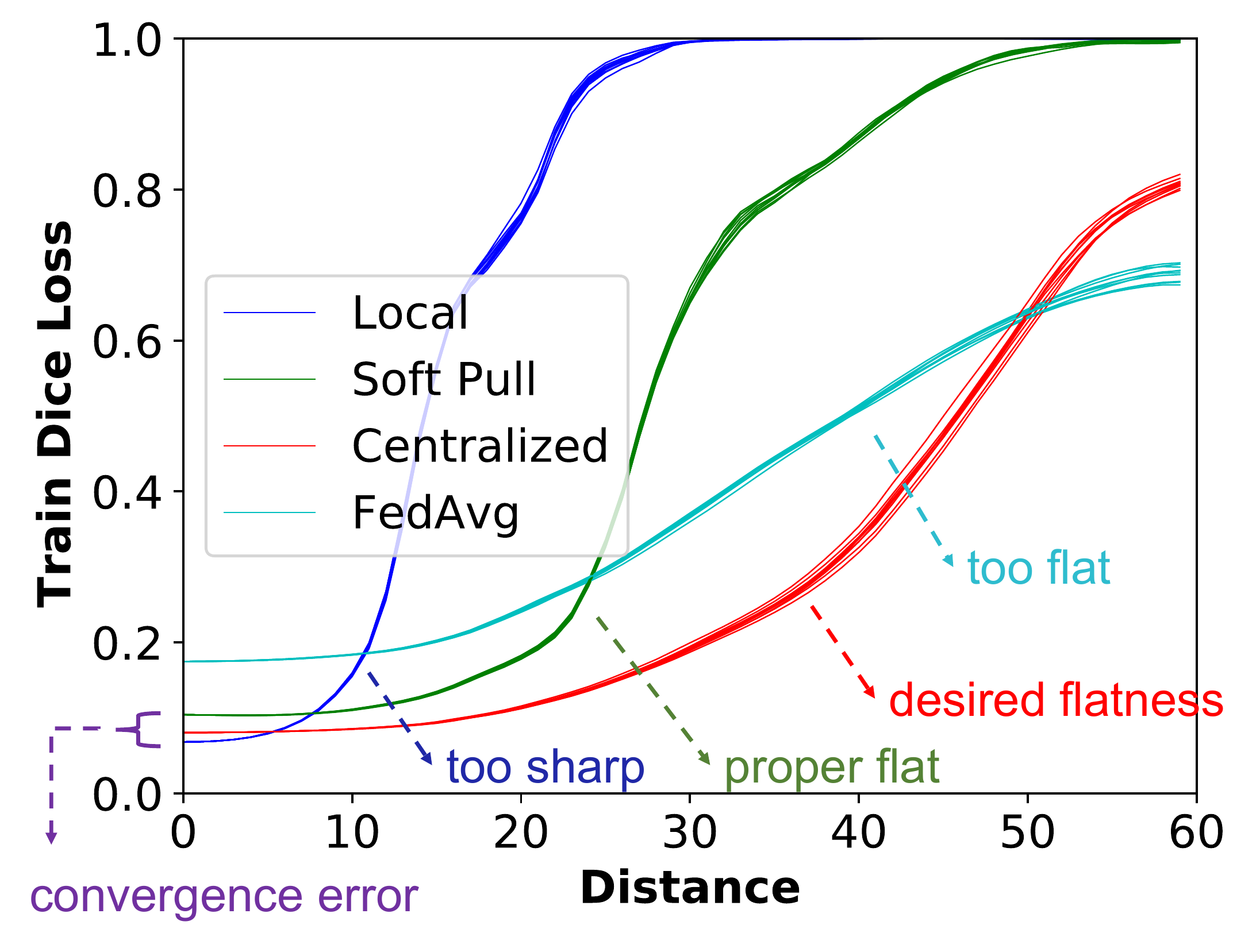}
    \includegraphics[width=0.3\textwidth]{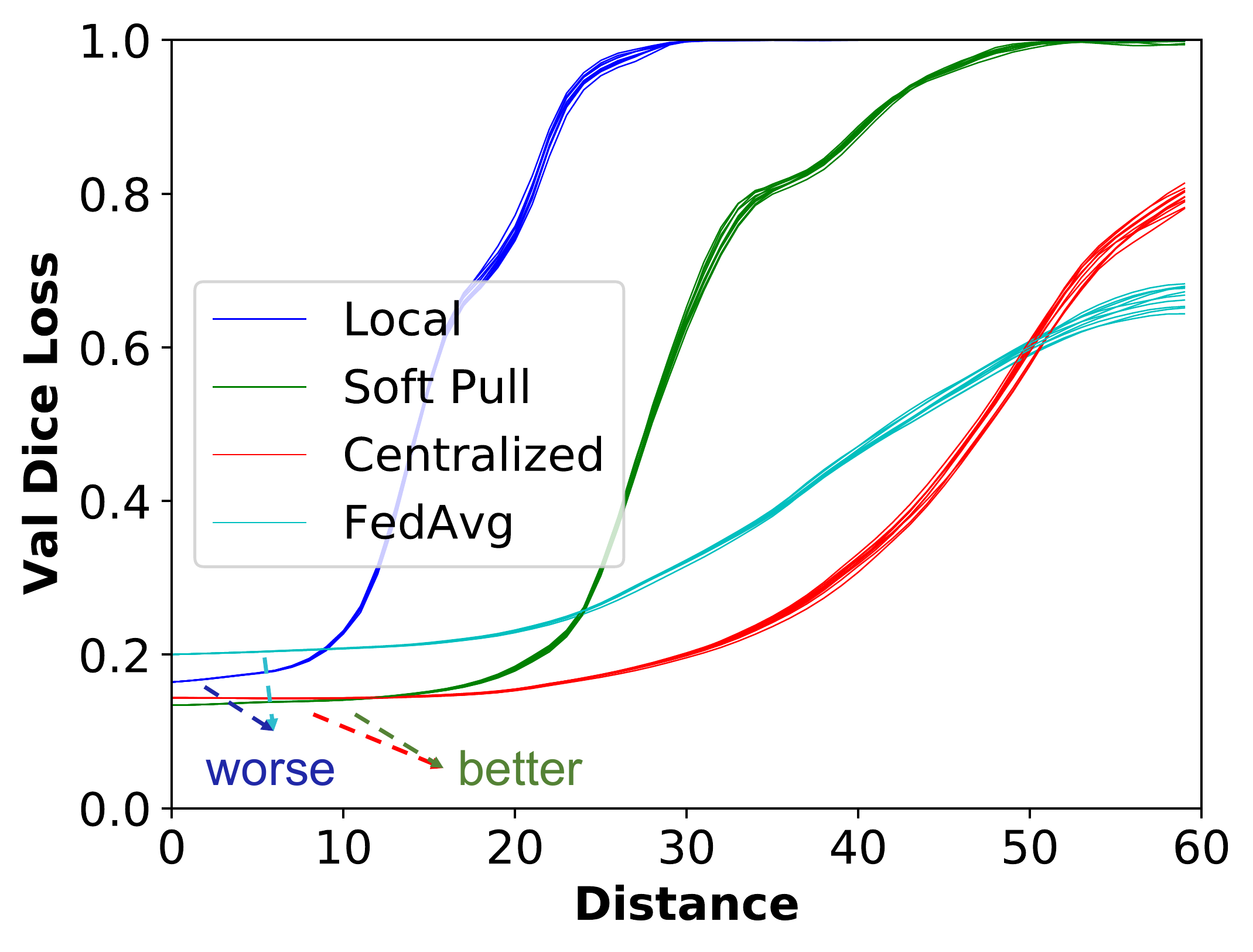}
    \includegraphics[width=0.3\textwidth]{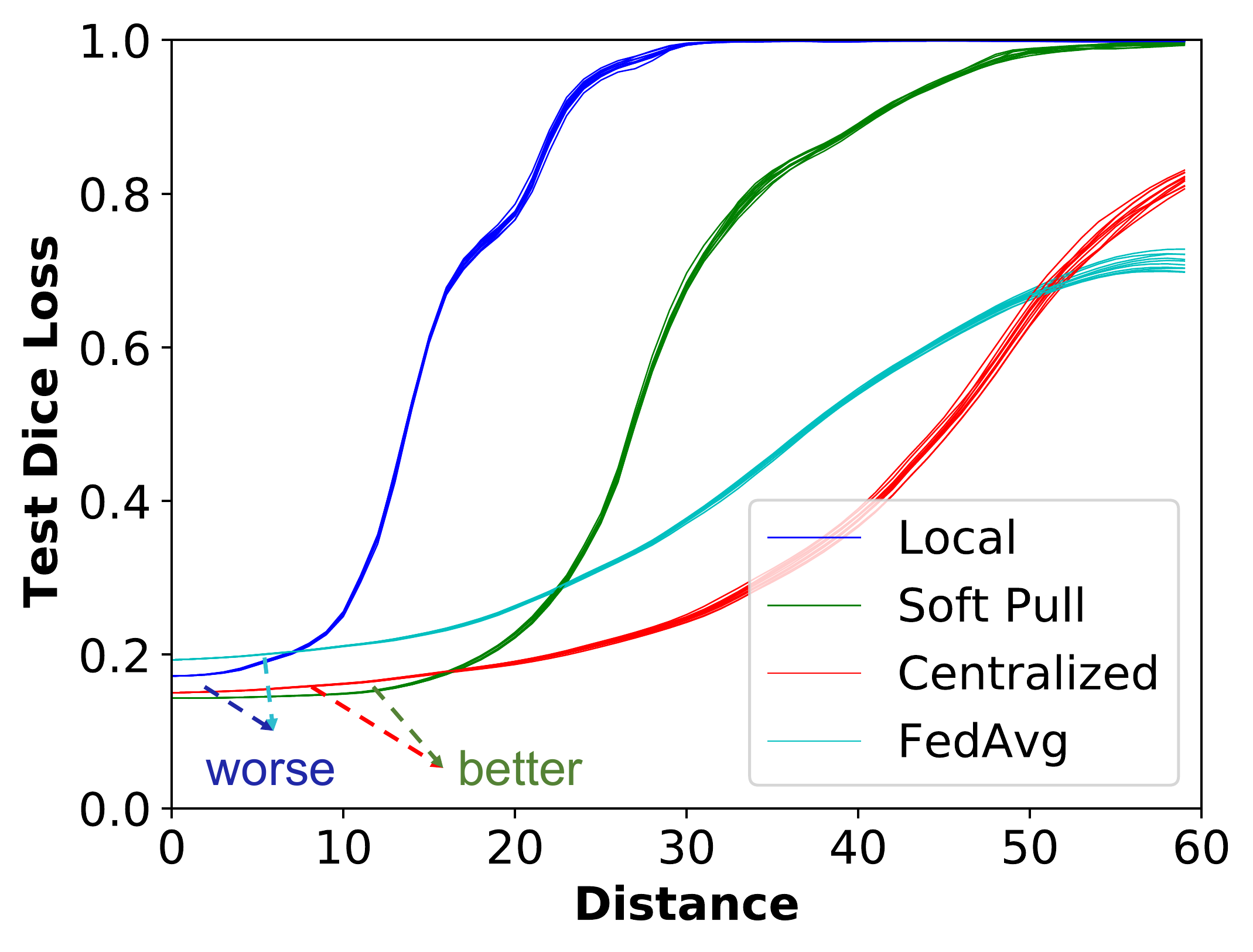}
    \caption{The 1D loss surface near the models trained by different methods on Client 5's data in retinal segmentation.}
    \label{fig:loss surface}
\end{figure*}

\begin{table*}[t]
\small
    \centering
    \begin{tabular}{c|cccccc|cc}
        \toprule
        Method & Client 1 & Client 2 & Client 3 & Client 4 & Client 5 & Client 6 & Client Avg Dice & Global Dice \\
        \midrule
        FT \cite{wang2019federated} & 0.9087 & 0.8703 & 0.8877 & 0.9003 & 0.8409 & 0.9151 & 0.8875 & 0.8984 \\
        APFL \cite{deng2020adaptive} & 0.9083 & 0.8640 & 0.8794 & 0.8969 & 0.8416 & 0.9152 & 0.8842 & 0.8966 \\
        Per-FedAvg \cite{fallah2020personalized} & 0.9051 & 0.8559 & 0.8708 & 0.8954 & 0.8031 & 0.9119 & 0.8737 & 0.8900 \\
        Per-FedMe \cite{t2020personalized} & 0.9084 & 0.8646 & 0.8822 & 0.8980 & 0.8211 & 0.9162 & 0.8818 & 0.8957 \\
        SoftPull & \textbf{0.9132} & \textbf{0.8769} & \textbf{0.8865} & \textbf{0.9041} & \textbf{0.8483} & \textbf{0.9195} & \textbf{0.8914} & \textbf{0.9028} \\
        \bottomrule
    \end{tabular}
    \caption{FedSM with different personalization method in retinal segmentation. Dice = average of disc and cup Dice coefficients.}
    \label{tab:vary personalization}
\end{table*}

\begin{table}[t]
\small
    \centering
    \begin{tabular}{c|ccccc}
        \toprule
        $\lambda$ & 0.1 & 0.3 & 0.5 & \textbf{0.7} & 0.9 \\
        \midrule
        Client Avg & 0.8808 & 0.8859 & 0.8895 & \textbf{0.8914} & 0.8882 \\
        Global & 0.8964 & 0.8896 & 0.9019 & \textbf{0.9028} & 0.9001 \\
        \bottomrule
    \end{tabular}
    \caption{FedSM with different coefficient $\lambda$ in retinal segmentation. Dice = average of disc and cup Dice coefficients.}
    \label{tab:vary lambda}
    \vspace{-13pt}
\end{table}

We validate our proposed method on three real-world FL medical image segmentation tasks: retinal disc \& cup from 2D fundus images, and prostate segmentation from 3D MR images. The global and personalized model architecture is 2D U-Net \cite{ronneberger2015u}, while the model selector architecture is VGG-11 \cite{simonyan2014very}. We randomly split the data to train/validation/test with a ratio of 0.5/0.25/0.25. The image data are resized to $256\times 256$. The local training epoch is 1 and the total training rounds is 150. Most methods converge in 100 rounds. But for FedSM-extra, we train the global and personalized models for 100 rounds and the model selector for an extra 50 rounds. The loss function is Dice loss and the test metric is Dice coefficient. The base optimizer is Adam with $\beta=(0.9, 0.999)$. We tune the best learning rate for all methods and the threshold $\gamma$ for FedSM. For prostate segmentation, in particular, the image data are 3D but we take the 2D slices and perform 2D segmentation. Each experiment repeatedly runs 3 times and we report the mean value.

The dataset information is summarized in Table \ref{tab:retinal num} and \ref{tab:prostate num 2d}. Overall, the retinal dataset features lower data similarity among clients (stronger non-iid). The images may differ in position, color, brightness, background ratio, etc. While the prostate dataset has a higher data similarity as the images mostly differ in brightness (see Appendix \ref{appendix:additional dataset info}).

We compare FedSM and FedSM-extra with baselines (1) Centralized: centralized training, which is the upper bound but prohibited in FL, (2) Local: local training on one client, (3) FedAvg \cite{mcmahan2017communication}, the \textit{de facto} FL method, (4) FedProx \cite{li2018federated}, and (5) Scaffold \cite{karimireddy2020scaffold}.

\subsection{General Results}

We compare the training curves of different methods in Figure \ref{fig:curve}. The centralized training upper bound is plotted as a horizontal dash line. We can see that the proposed FedSM is the only FL method to close the validation gap to centralized training. FedSM is even better than centralized training on the retinal cup segmentation task, due to the proposed SoftPull personalization method. Note that we can not show the training curve of FedSM-extra as its model selector has to be trained in the extra training rounds.

We summarize the testing numbers in Table \ref{tab:retinal avg dice} and \ref{tab:prostate dice}. For retinal segmentation, FedSM slightly improves centralized training regarding the client average Dice and global Dice by 0.2\% and 0.1\% respectively, while FedAvg shows a decrease of 1.9\% and 0.9\%. The FedSM-extra shows the same performance as FedSM, validating the proposed simplification from Eq.~(\ref{eq:fedsm-extra}) to Eq.~(\ref{eq:fedsm}). For prostate segmentation, similar patterns can be observed. But the gap becomes smaller due to higher data similarity among clients.

For retinal segmentation, FedSM outperforms centralized training for client 3 and matches centralized training for the other clients. However, FedAvg is inferior to centralized training for clients 1, 2, 3, and 5 where the local dataset size is smaller. What's more, FedAvg shows similar test Dice performance to local training for clients 1 and 2, and is even inferior to local training for clients 3 and 5. Therefore, those clients do not benefit from FL via FedAvg, and may not be willing to join the FL system.

We also observe that local training does not generalize well on other clients' data, which is critical as it will perform poorly for patients from other clients (medical institutions). Centralized training improves the local training on the local dataset, especially for clients with insufficient data.

\subsection{Validate Motivation}

\textbf{Validate FedSM.} Recall that our first motivation is to find the closest local data distribution for the test data. In FedSM, we first plot the TSNE map of the features extracted from the model selector in Figure \ref{fig:tsne}. To validate that the model selector can fulfill our motivation, we sequentially choose client $k\in\{1,2,\cdots,6\}$ as the unseen client to test and FL train the model with clients $\{1,2,\cdots,6\}/\{k\}$. We set the threshold $\gamma=0$ to let the model selector select from the personalized models. We summarize the frequency in Table \ref{tab:model selector frequency}. We can see that the model selector tends to select the personalized models of clients 3 and 5 for client 6, which also matches Figure \ref{fig:tsne} and the local training results in Table \ref{tab:retinal avg dice} that clients 3 and 5 are more similar to client 6. Similar patterns can be observed for the other clients. Therefore, the model selector indeed fulfills our motivation. Note that to validate the model selector, we cannot let the unseen client $k$ join the FL system. Because in that case, the model selector tends to select its own personalized model.

In Table \ref{tab:model selector frequency}, we also validate that the threshold $\gamma$ helps improve the performance of FedSM for the unseen data. For those unseen data with low confidence from the model selector, a larger $\gamma$ increases the chance of the global model to be selected because maybe none of the personalized models is suitable. By choosing a proper $\gamma$, we can further improve the Dice of unseen clients 5 and 6 by 3\%.

\textbf{Validate SoftPull.} Recall that our second motivation is to find a model generalizing well on the local data distribution even with insufficient local data. To achieve it we propose a new personalized FL optimization formulation with SoftPull to solve it. The Remark \ref{remark:sim} of the theoretical analysis can be empirically validated by the fact that the best $\lambda=0.7$ (closer to 1) for the retinal segmentation task with lower data similarity, and that the best $\lambda=0.3$ (closer to $\frac{1}{K}=\frac{1}{6}=0.17$) for the prostate segmentation task with higher data similarity.

Next, we will validate Remark \ref{remark:error} that a proper $\lambda$ may lead to a convergence error, but in the meantime may improve the generalization by preventing overfitting the small local dataset with the help of other clients. We plot the 1D loss surface near the trained model by computing the loss along 10 randomly sampled unit vector directions (Figure \ref{fig:loss surface}), following existing works \cite{izmailov2018averaging,he2019asymmetric}. It is interesting to see that local training overfits the training data and leads to a sharp local training optimum, which is known to generalize worse \cite{izmailov2018averaging,yang2019swalp,he2019asymmetric}. On the contrary, we observe an ``over-regularization'' effect for FedAvg as it has an even flatter training optimum than centralized training and a large convergence error (worse training loss), which also leads to a worse generalization performance. Indeed, averaging model in FedAvg can be regarded as a sort of implicit regularization. In comparison, SoftPull achieves a tunable flatness by choosing a proper $\lambda$. Even if it leads to a convergence error, it achieves generalization performance better than local training and comparable to centralized training.

\subsection{Ablation Study}
\textbf{Personalization.} We compare personalization methods in FedSM in Table \ref{tab:vary personalization}, including (1) FT (local fine-tuning) \cite{wang2019federated}, (2) APFL \cite{deng2020adaptive}, (3) Per-FedAvg \cite{fallah2020personalized}, and (4) Per-FedMe \cite{t2020personalized}. All methods' hyper-parameters are tuned for best results. SoftPull is the better interpolation method among them, outperforming APFL by 0.62\% regarding the global Dice coefficient. It also outperforms the best counterpart by 0.44\%.

\textbf{Interpolation Coefficient $\lambda$.} We explore different $\lambda$ values of FedSM in Table \ref{tab:vary lambda} and $\lambda=0.7$ performs the best.

\section{Conclusion}
In this work, we propose FedSM to close the generalization gap between FL and centralized training for medical image segmentation for the first time. The empirical study on real-world medical FL tasks validates our theoretical analysis and motivation to avoid the client drift issue.

\clearpage
{\small
\bibliographystyle{ieee_fullname}
\bibliography{egbib}

\begin{thebibliography}{10}\itemsep=-1pt

\bibitem{nci-isbi}
Nci\_isbi dataset.
\newblock https://www.cancerimagingarchive.net/.

\bibitem{promise12}
Promise12 dataset.
\newblock https://promise12.grand-challenge.org/.

\bibitem{prostatex}
Prostatex dataset.
\newblock https://prostatex.grand-challenge.org/.

\bibitem{refuge}
Refuge dataset.
\newblock https://refuge.grand-challenge.org/details/.

\bibitem{acar2020federated}
Durmus Alp~Emre Acar, Yue Zhao, Ramon Matas, Matthew Mattina, Paul Whatmough,
  and Venkatesh Saligrama.
\newblock Federated learning based on dynamic regularization.
\newblock In {\em International Conference on Learning Representations}, 2020.

\bibitem{almazroa2018retinal}
Ahmed Almazroa, Sami Alodhayb, Essameldin Osman, Eslam Ramadan, Mohammed
  Hummadi, Mohammed Dlaim, Muhannad Alkatee, Kaamran Raahemifar, and Vasudevan
  Lakshminarayanan.
\newblock Retinal fundus images for glaucoma analysis: the riga dataset.
\newblock In {\em Medical Imaging 2018: Imaging Informatics for Healthcare,
  Research, and Applications}, volume 10579, page 105790B. International
  Society for Optics and Photonics, 2018.

\bibitem{antonelli2021medical}
Michela Antonelli, Annika Reinke, Spyridon Bakas, Keyvan Farahani, Bennett~A
  Landman, Geert Litjens, Bjoern Menze, Olaf Ronneberger, Ronald~M Summers,
  Bram van Ginneken, et~al.
\newblock The medical segmentation decathlon.
\newblock {\em arXiv preprint arXiv:2106.05735}, 2021.

\bibitem{defazio2014saga}
Aaron Defazio, Francis Bach, and Simon Lacoste-Julien.
\newblock Saga: A fast incremental gradient method with support for
  non-strongly convex composite objectives.
\newblock {\em Advances in neural information processing systems},
  27:1646--1654, 2014.

\bibitem{defazio2018ineffectiveness}
Aaron Defazio and L{\'e}on Bottou.
\newblock On the ineffectiveness of variance reduced optimization for deep
  learning.
\newblock {\em arXiv preprint arXiv:1812.04529}, 2018.

\bibitem{deng2020adaptive}
Yuyang Deng, Mohammad~Mahdi Kamani, and Mehrdad Mahdavi.
\newblock Adaptive personalized federated learning.
\newblock {\em arXiv preprint arXiv:2003.13461}, 2020.

\bibitem{fallah2020personalized}
Alireza Fallah, Aryan Mokhtari, and Asuman Ozdaglar.
\newblock Personalized federated learning with theoretical guarantees: A
  model-agnostic meta-learning approach.
\newblock {\em Advances in Neural Information Processing Systems}, 33, 2020.

\bibitem{fumero2011rim}
Francisco Fumero, Silvia Alay{\'o}n, Jos{\'e}~L Sanchez, Jose Sigut, and M
  Gonzalez-Hernandez.
\newblock Rim-one: An open retinal image database for optic nerve evaluation.
\newblock In {\em 2011 24th international symposium on computer-based medical
  systems (CBMS)}, pages 1--6. IEEE, 2011.

\bibitem{gao2021convergence}
Hongchang Gao, An Xu, and Heng Huang.
\newblock On the convergence of communication-efficient local sgd for federated
  learning.
\newblock In {\em Proceedings of the AAAI Conference on Artificial
  Intelligence, Virtual}, pages 18--19, 2021.

\bibitem{geiping2020inverting}
Jonas Geiping, Hartmut Bauermeister, Hannah Dr{\"o}ge, and Michael Moeller.
\newblock Inverting gradients--how easy is it to break privacy in federated
  learning?
\newblock {\em arXiv preprint arXiv:2003.14053}, 2020.

\bibitem{ghosh2020efficient}
Avishek Ghosh, Jichan Chung, Dong Yin, and Kannan Ramchandran.
\newblock An efficient framework for clustered federated learning.
\newblock {\em Advances in Neural Information Processing Systems}, 33, 2020.

\bibitem{gu2021privacy}
Bin Gu, An Xu, Zhouyuan Huo, Cheng Deng, and Heng Huang.
\newblock Privacy-preserving asynchronous vertical federated learning
  algorithms for multiparty collaborative learning.
\newblock {\em IEEE Transactions on Neural Networks and Learning Systems},
  2021.

\bibitem{guo2021multi}
Pengfei Guo, Puyang Wang, Jinyuan Zhou, Shanshan Jiang, and Vishal~M Patel.
\newblock Multi-institutional collaborations for improving deep learning-based
  magnetic resonance image reconstruction using federated learning.
\newblock In {\em Proceedings of the IEEE/CVF Conference on Computer Vision and
  Pattern Recognition}, pages 2423--2432, 2021.

\bibitem{he2019asymmetric}
Haowei He, Gao Huang, and Yang Yuan.
\newblock Asymmetric valleys: Beyond sharp and flat local minima.
\newblock {\em arXiv preprint arXiv:1902.00744}, 2019.

\bibitem{he2016deep}
Kaiming He, Xiangyu Zhang, Shaoqing Ren, and Jian Sun.
\newblock Deep residual learning for image recognition.
\newblock In {\em Proceedings of the IEEE conference on computer vision and
  pattern recognition}, pages 770--778, 2016.

\bibitem{hsieh2020non}
Kevin Hsieh, Amar Phanishayee, Onur Mutlu, and Phillip Gibbons.
\newblock The non-iid data quagmire of decentralized machine learning.
\newblock In {\em International Conference on Machine Learning}, pages
  4387--4398. PMLR, 2020.

\bibitem{hsu2019measuring}
Tzu-Ming~Harry Hsu, Hang Qi, and Matthew Brown.
\newblock Measuring the effects of non-identical data distribution for
  federated visual classification.
\newblock {\em arXiv preprint arXiv:1909.06335}, 2019.

\bibitem{ioffe2015batch}
Sergey Ioffe and Christian Szegedy.
\newblock Batch normalization: Accelerating deep network training by reducing
  internal covariate shift.
\newblock In {\em International conference on machine learning}, pages
  448--456. PMLR, 2015.

\bibitem{izmailov2018averaging}
Pavel Izmailov, Dmitrii Podoprikhin, Timur Garipov, Dmitry Vetrov, and
  Andrew~Gordon Wilson.
\newblock Averaging weights leads to wider optima and better generalization.
\newblock {\em arXiv preprint arXiv:1803.05407}, 2018.

\bibitem{jiang2019improving}
Yihan Jiang, Jakub Kone{\v{c}}n{\`y}, Keith Rush, and Sreeram Kannan.
\newblock Improving federated learning personalization via model agnostic meta
  learning.
\newblock {\em arXiv preprint arXiv:1909.12488}, 2019.

\bibitem{johnson2013accelerating}
Rie Johnson and Tong Zhang.
\newblock Accelerating stochastic gradient descent using predictive variance
  reduction.
\newblock {\em Advances in neural information processing systems}, 26:315--323,
  2013.

\bibitem{kairouz2019advances}
Peter Kairouz, H~Brendan McMahan, Brendan Avent, Aur{\'e}lien Bellet, Mehdi
  Bennis, Arjun~Nitin Bhagoji, Kallista Bonawitz, Zachary Charles, Graham
  Cormode, Rachel Cummings, et~al.
\newblock Advances and open problems in federated learning.
\newblock {\em arXiv preprint arXiv:1912.04977}, 2019.

\bibitem{karimireddy2020scaffold}
Sai~Praneeth Karimireddy, Satyen Kale, Mehryar Mohri, Sashank Reddi, Sebastian
  Stich, and Ananda~Theertha Suresh.
\newblock Scaffold: Stochastic controlled averaging for federated learning.
\newblock In {\em International Conference on Machine Learning}, pages
  5132--5143. PMLR, 2020.

\bibitem{kingma2014adam}
Diederik~P Kingma and Jimmy Ba.
\newblock Adam: A method for stochastic optimization.
\newblock {\em arXiv preprint arXiv:1412.6980}, 2014.

\bibitem{konevcny2016federated}
Jakub Kone{\v{c}}n{\`y}, H~Brendan McMahan, Felix~X Yu, Peter Richt{\'a}rik,
  Ananda~Theertha Suresh, and Dave Bacon.
\newblock Federated learning: Strategies for improving communication
  efficiency.
\newblock {\em arXiv preprint arXiv:1610.05492}, 2016.

\bibitem{lemaitre2015computer}
Guillaume Lema{\^\i}tre, Robert Mart{\'\i}, Jordi Freixenet, Joan~C Vilanova,
  Paul~M Walker, and Fabrice Meriaudeau.
\newblock Computer-aided detection and diagnosis for prostate cancer based on
  mono and multi-parametric mri: a review.
\newblock {\em Computers in biology and medicine}, 60:8--31, 2015.

\bibitem{li2014scaling}
Mu Li, David~G Andersen, Jun~Woo Park, Alexander~J Smola, Amr Ahmed, Vanja
  Josifovski, James Long, Eugene~J Shekita, and Bor-Yiing Su.
\newblock Scaling distributed machine learning with the parameter server.
\newblock In {\em 11th $\{$USENIX$\}$ Symposium on Operating Systems Design and
  Implementation ($\{$OSDI$\}$ 14)}, pages 583--598, 2014.

\bibitem{li2021ditto}
Tian Li, Shengyuan Hu, Ahmad Beirami, and Virginia Smith.
\newblock Ditto: Fair and robust federated learning through personalization.
\newblock In {\em International Conference on Machine Learning}, pages
  6357--6368. PMLR, 2021.

\bibitem{li2018federated}
Tian Li, Anit~Kumar Sahu, Manzil Zaheer, Maziar Sanjabi, Ameet Talwalkar, and
  Virginia Smith.
\newblock Federated optimization in heterogeneous networks.
\newblock {\em arXiv preprint arXiv:1812.06127}, 2018.

\bibitem{Li2020Fair}
Tian Li, Maziar Sanjabi, Ahmad Beirami, and Virginia Smith.
\newblock Fair resource allocation in federated learning.
\newblock In {\em International Conference on Learning Representations}, 2020.

\bibitem{liang2019variance}
Xianfeng Liang, Shuheng Shen, Jingchang Liu, Zhen Pan, Enhong Chen, and Yifei
  Cheng.
\newblock Variance reduced local sgd with lower communication complexity.
\newblock {\em arXiv preprint arXiv:1912.12844}, 2019.

\bibitem{liu2021feddg}
Quande Liu, Cheng Chen, Jing Qin, Qi Dou, and Pheng-Ann Heng.
\newblock Feddg: Federated domain generalization on medical image segmentation
  via episodic learning in continuous frequency space.
\newblock In {\em Proceedings of the IEEE/CVF Conference on Computer Vision and
  Pattern Recognition}, pages 1013--1023, 2021.

\bibitem{mansour2020three}
Yishay Mansour, Mehryar Mohri, Jae Ro, and Ananda~Theertha Suresh.
\newblock Three approaches for personalization with applications to federated
  learning.
\newblock {\em arXiv preprint arXiv:2002.10619}, 2020.

\bibitem{mcmahan2017communication}
Brendan McMahan, Eider Moore, Daniel Ramage, Seth Hampson, and Blaise~Aguera y
  Arcas.
\newblock Communication-efficient learning of deep networks from decentralized
  data.
\newblock In {\em Artificial Intelligence and Statistics}, pages 1273--1282.
  PMLR, 2017.

\bibitem{mohri2019agnostic}
Mehryar Mohri, Gary Sivek, and Ananda~Theertha Suresh.
\newblock Agnostic federated learning.
\newblock In {\em International Conference on Machine Learning}, pages
  4615--4625, 2019.

\bibitem{reddi2020adaptive}
Sashank~J Reddi, Zachary Charles, Manzil Zaheer, Zachary Garrett, Keith Rush,
  Jakub Kone{\v{c}}n{\`y}, Sanjiv Kumar, and Hugh~Brendan McMahan.
\newblock Adaptive federated optimization.
\newblock In {\em International Conference on Learning Representations}, 2020.

\bibitem{ronneberger2015u}
Olaf Ronneberger, Philipp Fischer, and Thomas Brox.
\newblock U-net: Convolutional networks for biomedical image segmentation.
\newblock In {\em International Conference on Medical image computing and
  computer-assisted intervention}, pages 234--241. Springer, 2015.

\bibitem{simonyan2014very}
Karen Simonyan and Andrew Zisserman.
\newblock Very deep convolutional networks for large-scale image recognition.
\newblock {\em arXiv preprint arXiv:1409.1556}, 2014.

\bibitem{6867807}
J. {Sivaswamy}, S.~R. {Krishnadas}, G. {Datt Joshi}, M. {Jain}, and A.~U. {Syed
  Tabish}.
\newblock Drishti-gs: Retinal image dataset for optic nerve head(onh)
  segmentation.
\newblock In {\em 2014 IEEE 11th International Symposium on Biomedical Imaging
  (ISBI)}, pages 53--56, April 2014.

\bibitem{srivastava2014dropout}
Nitish Srivastava, Geoffrey Hinton, Alex Krizhevsky, Ilya Sutskever, and Ruslan
  Salakhutdinov.
\newblock Dropout: a simple way to prevent neural networks from overfitting.
\newblock {\em The journal of machine learning research}, 15(1):1929--1958,
  2014.

\bibitem{t2020personalized}
Canh T~Dinh, Nguyen Tran, and Tuan~Dung Nguyen.
\newblock Personalized federated learning with moreau envelopes.
\newblock {\em Advances in Neural Information Processing Systems}, 33, 2020.

\bibitem{wang2020tackling}
Jianyu Wang, Qinghua Liu, Hao Liang, Gauri Joshi, and H.~Vincent Poor.
\newblock Tackling the objective inconsistency problem in heterogeneous
  federated optimization.
\newblock In {\em Advances in Neural Information Processing Systems},
  volume~33, pages 7611--7623, 2020.

\bibitem{wang2019federated}
Kangkang Wang, Rajiv Mathews, Chlo{\'e} Kiddon, Hubert Eichner, Fran{\c{c}}oise
  Beaufays, and Daniel Ramage.
\newblock Federated evaluation of on-device personalization.
\newblock {\em arXiv preprint arXiv:1910.10252}, 2019.

\bibitem{xu2021double}
An Xu and Heng Huang.
\newblock Double momentum sgd for federated learning.
\newblock {\em arXiv preprint arXiv:2102.03970}, 2021.

\bibitem{xu2020acceleration}
An Xu, Zhouyuan Huo, and Heng Huang.
\newblock On the acceleration of deep learning model parallelism with
  staleness.
\newblock In {\em Proceedings of the IEEE/CVF Conference on Computer Vision and
  Pattern Recognition}, pages 2088--2097, 2020.

\bibitem{xu2021step}
An Xu, Zhouyuan Huo, and Heng Huang.
\newblock Step-ahead error feedback for distributed training with compressed
  gradient.
\newblock In {\em Proceedings of the AAAI Conference on Artificial
  Intelligence}, volume~35, pages 10478--10486, 2021.

\bibitem{yang2019swalp}
Guandao Yang, Tianyi Zhang, Polina Kirichenko, Junwen Bai, Andrew~Gordon
  Wilson, and Chris De~Sa.
\newblock Swalp: Stochastic weight averaging in low precision training.
\newblock In {\em International Conference on Machine Learning}, pages
  7015--7024. PMLR, 2019.

\bibitem{yin2021see}
Hongxu Yin, Arun Mallya, Arash Vahdat, Jose~M Alvarez, Jan Kautz, and Pavlo
  Molchanov.
\newblock See through gradients: Image batch recovery via gradinversion.
\newblock In {\em Proceedings of the IEEE/CVF Conference on Computer Vision and
  Pattern Recognition}, pages 16337--16346, 2021.

\bibitem{yu2019linear}
Hao Yu, Rong Jin, and Sen Yang.
\newblock On the linear speedup analysis of communication efficient momentum
  sgd for distributed non-convex optimization.
\newblock In {\em International Conference on Machine Learning}, pages
  7184--7193. PMLR, 2019.

\bibitem{zhao2020idlg}
Bo Zhao, Konda~Reddy Mopuri, and Hakan Bilen.
\newblock idlg: Improved deep leakage from gradients.
\newblock {\em arXiv preprint arXiv:2001.02610}, 2020.

\bibitem{zhu2019deep}
Ligeng Zhu, Zhijian Liu, and Song Han.
\newblock Deep leakage from gradients.
\newblock {\em Advances in Neural Information Processing Systems},
  32:14774--14784, 2019.

\end{thebibliography}
}

\clearpage
\appendix
\section{Additional Dataset Information}\label{appendix:additional dataset info}
\begin{table}[htb!]
\small
    \centering
    \begin{tabular}{c|cccccc|c}
        \toprule
        Client & 1 & 2 & 3 & 4 & 5 & 6 & Global \\
        \midrule
        Train & 10 & 16 & 18 & 18 & 25 & 50 & 137 \\
        Val & 5 & 8 & 9 & 9 & 12 & 25 & 68 \\
        Test & 4 & 8 & 8 & 8 & 13 & 24 & 65 \\
        \bottomrule
    \end{tabular}
    \caption{Prostate dataset: number of data (3D image) in each client.}
    \label{tab:prostate num 3d}
\end{table}
\begin{figure}[htb!]
    \centering
    \includegraphics[width=0.32\columnwidth]{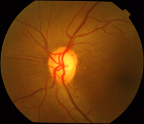}
    \includegraphics[width=0.32\columnwidth]{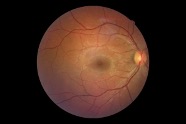}
    \includegraphics[width=0.32\columnwidth]{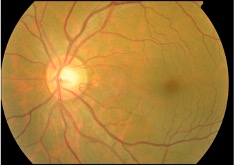}
    \includegraphics[width=0.32\columnwidth]{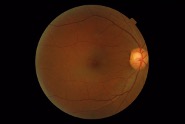}
    \includegraphics[width=0.32\columnwidth]{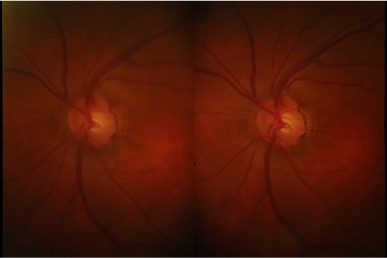}
    \includegraphics[width=0.32\columnwidth]{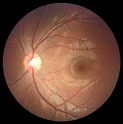}
    \caption{Representative original 2D image in retinal dataset (low data similarity). First row: client 1 to 3. Second row: client 4 to 6.}
    \label{fig:retinal original image}
\end{figure}
\begin{figure}[htb!]
    \centering
    \includegraphics[width=0.32\columnwidth]{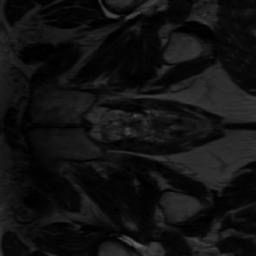}
    \includegraphics[width=0.32\columnwidth]{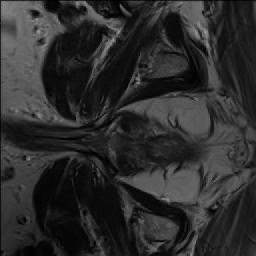}
    \includegraphics[width=0.32\columnwidth]{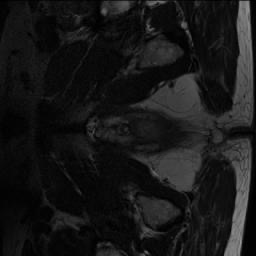}
    \includegraphics[width=0.32\columnwidth]{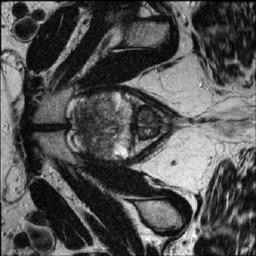}
    \includegraphics[width=0.32\columnwidth]{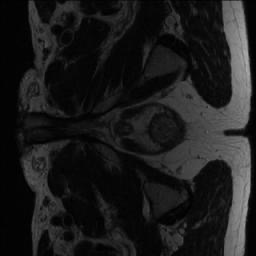}
    \includegraphics[width=0.32\columnwidth]{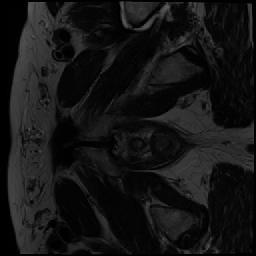}
    \caption{Representative original 2D image slices in prostate dataset (high data similarity). First row: client 1 to 3. Second row: client 4 to 6. E.g., the first slice comes from a 3D image in client 1.}
    \label{fig:prostate original image}
\end{figure}

\section{FedSM-extra Algorithm}\label{appendix: fedsm-extra}

\begin{algorithm}[htb!]
\caption{FedSM-extra training.}\label{alg:fedsm-extra training}
\begin{algorithmic}[1]
    \STATE \textbf{Input:} local dataset $\mathcal{D}_k$, rounds $R$, number of sites $K$, learning rate $\eta$, coefficient $\lambda$, client weight $\frac{n_k}{n}$.
    \STATE \textbf{Initialize:} global model $w_g$, personalized model $w_{p,k}$, model selector $w_s$, base optimizer $\text{OPT}(\cdot)$
    
    \FOR{round $r=1,2,\cdots,R$}
        \STATE SERVER: send models ($w_g$, $w_{p,k}$) to client $k$.
        \FOR{CLIENT $k\in\{1,2,\cdots,K\}$ in parallel}
            \STATE initialize $w_{g,k}\leftarrow w_g$
            \FOR{batch $(x,y) \in \mathcal{D}_n$}
                \STATE $w_{g,k}\leftarrow \text{OPT}(w_{g,k},\eta,\nabla_{w_{g,k}} L(f(w_{g,k};x),y))$
                \STATE $w_{p,k}\leftarrow \text{OPT}(w_{p,k},\eta,\nabla_{w_{p,k}} L(f(w_{p,k};x),y))$
            \ENDFOR
            \STATE send ($w_{g,k}$, $w_{p,k}$) to server
        \ENDFOR
        \STATE SERVER: $w_g\leftarrow \sum^{K}_{k=1}\frac{n_k}{n} w_{g,k}$ \hfill // \textit{FedAvg}
        \STATE SERVER: $\forall k \in \{1,2,\cdots,K\}$, $w_{p,k}\leftarrow \lambda w_{p,k} + (1-\lambda)\frac{1}{K-1}\sum^{K}_{k^\prime=1, k^\prime\neq k} w_{p,k^\prime}$ \hfill // \textit{SoftPull}
    \ENDFOR
    
    \STATE // \textit{extra training rounds}
    \STATE SERVER: send models ($w_g$, $w_{p,n}$) to clients.
    \FOR{round $r=1,2,\cdots,\Delta R$}
        \STATE SERVER: send model $w_s$ to clients.
        \FOR{CLIENT $k\in\{1,2,\cdots,K\}$ in parallel}
            \STATE Initialize $w_{s,k}\leftarrow w_s$
            \FOR{batch $(x,y) \in \mathcal{D}_n$}
                \STATE // \textit{$y_s$ from Eq.~(\ref{eq:fedsm-extra}})
                \STATE $w_{s,k}\leftarrow \text{OPT}(w_{s,k}, \eta_s,\nabla_{w_{s,k}}L_s(f_s(w_{s,k};x),y_s))$
            \ENDFOR
            \STATE send $w_{s,k}$ to server
        \ENDFOR
        \STATE SERVER: $w_s\leftarrow \sum^{K}_{k=1}\frac{n_k}{n} w_{s,k}$
    \ENDFOR
    
    \STATE \textbf{Output:} model ($w_g$, $\{w_{p,k}\}^{K}_{k=1}$, $w_s$)
\end{algorithmic}
\end{algorithm}

\begin{algorithm}[htb!]
\caption{FedSM-extra inference.}\label{alg:fedsm-extra inference}
\begin{algorithmic}[1]
    \STATE \textbf{Input:} data $x$, model ($w_g$, $\{w_{p,k}\}^{K}_{k=1}$, $w_s$)
    
    \STATE $\widehat{y}_s = f_s(w_s; x)$
    \STATE $k=\arg\max(\widehat{y}_s)\in\{0,1,\cdots,K\}$
    
    \IF {$k>0$}
        \STATE $\widehat{y}=f(w_{p,k};x)$
    \ELSE
        \STATE $\widehat{y}=f(w_g;x)$
    \ENDIF
    
    \STATE \textbf{Output:} $\widehat{y}$
\end{algorithmic}
\end{algorithm}

For the training of FedSM-extra, we train the global model and personalized models first, and then train the model selector, which incurs extra $\Delta R$ training rounds. In each training round of FedSM-extra, the communication cost is $2 w_g$ in the previous $R$ rounds (the global and personalized models have the same model architecture). It becomes $w_s$ in the extra $\Delta R$ training rounds.

For the inference of FedSM-extra, both the global model and personalized models can be selected. Therefore $k\in\{0,1,\cdots,K\}$ (For FedSM, $k\in\{1,2,\cdots,K\}$).

\section{Proof of SoftPull Convergence}\label{appendix:convergence}

Let the current/total training rounds be $r/R$, current/total local training steps be $m/M$, the current/total global training step be $t/T$. We denote the personalized model during training as $w_{p,k}^{r,m}$. For simplicity we use $f_k$ to denote loss $L_{\mathcal{D}_k}$.

After the local training in the last training round $r-1$ finishes, we get model $w_{p,k}^{r-1,M}$ and want to
\begin{equation}
    \min \sum^{K}_{k=1}f_k(\frac{1}{\lambda}w_{p,k}^{r-1,M}-\frac{1-\lambda}{\lambda}\frac{1}{K-1}\sum^{K}_{k^\prime=1,k^\prime\neq k}w_{p,k^\prime}^{r-1,M})
\end{equation}
In the beginning of the current training round $r$, from Eq.~(\ref{eq:softpull}), we will have
\begin{equation}
    w_{p,k}^{r,0}=\lambda w_{p,k}^{r-1,M} + (1-\lambda)\frac{1}{K-1}\sum^{K}_{k^\prime=1,k^\prime\neq k}w_{p,k^\prime}^{r-1,M}
\end{equation}
In the current training round $r$, we consider two stages. The first stage is a transition from then end of training round $r-1$ to the start of the current training round $r$, while the second stage is the start to the end of current training round $r$. Let $\lambda^\prime=\frac{K\lambda-1}{K-1}$, $1-\lambda^\prime=
\frac{K}{K-1}(1-\lambda)$, then
\begin{equation}
    w_{p,k}^{r,0} = \lambda^\prime w^{r-1,M}_{p,k} + (1-\lambda^\prime)\overline{w}_{p,k}^{r-1,M}
\end{equation}
where the bar denotes an average over all clients $k\in\{1,2,\cdots,K\}$. It can be clearly seen that when the data distributions of clients are very similar, we set $\lambda=\frac{1}{K}$, $\lambda^\prime=0$, i.e., the ``hard averaging" in FedAvg. When the data distributions are not similar at all, we set $\lambda=1$, $\lambda^\prime=1$ to only do local training. In other circumstances, theoretically we should set $\lambda\in[\frac{1}{K},1]$, $\lambda^\prime\in[0,1]$ according to the data similarity.

\subsection{Difference}
Suppose the stochastic gradient at iteration $(r,m)$ is $\nabla f_k(w_{p,k}^{r,m},x_{p,k}^{r,m})$ and the expected gradient is $\nabla f_k(w_{p,k}^{r,m})=\mathbb{E}_{x^{r,m}_{p,k}\in\mathcal{D}_k}\nabla f_k(w_{p,k}^{r,m},x_{p,k}^{r,m})=\mathbb{E}\nabla f_k(w_{p,k}^{r,m},x_{p,k}^{r,m})$. We need to bound
\begin{equation}
\begin{split}
    &\|(w_{p,k}^{r+1,0}-w_{p,k}^{r,M})\|^2_2\\
    &= \|(1-\lambda)(w_{p,k}^{r,M}-\frac{1}{K-1}\sum^{K}_{k^\prime=1,k^\prime\neq k}w_{p,k^\prime}^{r,M})\|^2_2\\
    &= (1-\lambda)^2 \|w_{p,k}^{r,M}-\frac{1}{K-1}(K\overline{w}^{r,M}_{p,k}-w_{p,k}^{r,M})\|^2_2\\
    &=\frac{(1-\lambda)^2K^2}{(K-1)^2}\|w_{p,k}^{r,M}-\overline{w}^{r,M}_{p,k}\|^2_2
\end{split}
\end{equation}
where
\begin{equation}
\begin{split}
    &\mathbb{E}\|w_{p,k}^{r,M}-\overline{w}^{r,M}_{p,k}\|^2_2\\
    &= \eta^2 \mathbb{E}\|\sum^{r}_{r^\prime=0}(\lambda^\prime)^{r-r^\prime}\sum^{M-1}_{m=0}[\nabla f_k(w_{p,k}^{r^\prime,m},x_{p,k}^{r^\prime,m})\\
    &\quad - \overline{\nabla f_k}(w_{p,k}^{r^\prime,m},x_{p,k}^{r^\prime,m})]\|^2_2\\
    &\leq \eta^2(\sum^{r}_{r^\prime=0}(\lambda^\prime)^{r-r^\prime})^2 \mathbb{E}\| \sum^{r}_{r^\prime=0} \frac{(\lambda^\prime)^{r-r^\prime}}{\sum^{r}_{r^\prime=0}(\lambda^\prime)^{r-r^\prime}}\sum^{M-1}_{m=0}\\
    &\quad [\nabla f_k(w_{p,k}^{r^\prime,m},x_{p,k}^{r^\prime,m}) - \overline{\nabla f_k}(w_{p,k}^{r^\prime,m},x_{p,k}^{r^\prime,m})]\|^2_2\\
    &\leq \eta^2(\sum^{r}_{r^\prime=0}(\lambda^\prime)^{r-r^\prime})\sum^{r}_{r^\prime=0}(\lambda^\prime)^{r-r^\prime}\mathbb{E}\|\sum^{M-1}_{m=0}\\
    &\quad [\nabla f_k(w_{p,k}^{r^\prime,m},x_{p,k}^{r^\prime,m}) - \overline{\nabla f_k}(w_{p,k}^{r^\prime,m},x_{p,k}^{r^\prime,m})]\|^2_2\\
    &\leq M\eta^2(\sum^{r}_{r^\prime=0}(\lambda^\prime)^{r-r^\prime})\sum^{r}_{r^\prime=0}(\lambda^\prime)^{r-r^\prime}\sum^{M-1}_{m=0}\\
    &\quad \mathbb{E}\|\nabla f_k(w_{p,k}^{r^\prime,m},x_{p,k}^{r^\prime,m}) - \overline{\nabla f_k}(w_{p,k}^{r^\prime,m},x_{p,k}^{r^\prime,m})\|^2_2\\
    &\leq 2M^2(G^2+\sigma^2)\eta^2(\sum^{r}_{r^\prime=0}(\lambda^\prime)^{r-r^\prime})^2\\
    &\leq 2M^2(G^2+\sigma^2)\eta^2 [\frac{1-(\lambda^\prime)^{r+1}}{1-\lambda^\prime}]^2
\end{split}
\end{equation}
where $\mathbb{E}\|\nabla f_k(w_{p,k}^{r^\prime,m},x_{p,k}^{r^\prime,m})\|^2_2 \leq 2(G^2+\sigma^2)$ based on Assumptions \ref{bounded gradient} and \ref{bounded variance}. Then
\begin{equation}
\begin{split}
    &\mathbb{E}\|(w_{p,k}^{r+1,0}-w_{p,k}^{r,M})\|^2_2\\
    &\leq \frac{[1-(\lambda^\prime)^{r+1}]^2(1-\lambda)^2K^2}{(1-\lambda^\prime)^2(K-1)^2} 2M^2(G^2+\sigma^2)\eta^2\\
    &= [1-(\lambda^\prime)^{r+1}]^2 2M^2(G^2+\sigma^2)\eta^2
\end{split}
\end{equation}

\subsection{Local Objective}

Here we consider the local objective function to optimize. From $(r,0)$ to $(r,M)$, i.e. $m\in\{0,1,\cdots,M-1\}$, due to the Lipschitz smooth assumption we have
\begin{equation}
\begin{split}
    &f_k(w_{k,p}^{r,m+1}) - f_k(w_{k,p}^{r,m})\\
    &\leq \langle\nabla f_k(w_{k,p}^{r,m}), w_{k,p}^{r,m+1}-w_{k,p}^{r,m}\rangle + \frac{L}{2}\|w_{k,p}^{r,m+1}-w_{k,p}^{r,m}\|^2_2\\
    &=-\eta\langle\nabla f_k(w_{k,p}^{r,m}),\nabla f_k(w_{k,p}^{r,m},x^{r,m}_{k,p})\rangle\\
    &\quad + \frac{\eta^2L}{2}\|\nabla f_k(w_{k,p}^{r,m},x^{r,m}_{k,p})\|^2_2\\
    &=-\eta\langle\nabla f_k(w_{k,p}^{r,m}),\nabla f_k(w_{k,p}^{r,m},x^{r,m}_{k,p})\rangle\\
    &\quad + \frac{\eta^2L}{2}\|\nabla f_k(w_{k,p}^{r,m})\|^2_2 + \frac{\eta^2L\sigma^2}{2}
\end{split}
\end{equation}
Take the expectation and suppose $\eta\leq\frac{1}{L}$,
\begin{equation}
\begin{split}
    &\mathbb{E}[f_k(w_{k,p}^{r,m+1}) - f_k(w_{k,p}^{r,m})]\\
    &\leq -\eta(1-\frac{\eta L}{2})\mathbb{E}\|\nabla f_k(w_{k,p}^{r,m})\|^2_2 + \frac{\eta^2L\sigma^2}{2}\\
    &\leq -\frac{\eta}{2}\mathbb{E}\|\nabla f_k(w_{k,p}^{r,m})\|^2_2 + \frac{\eta^2L\sigma^2}{2}
\end{split}
\end{equation}
\begin{equation}
\begin{split}
    &\mathbb{E}\|\nabla f_k(w_{k,p}^{r,m})\|^2_2\\
    &\leq \frac{2}{\eta}\mathbb{E}[f_k(w_{k,p}^{r,m})-f_k(w_{k,p}^{r,m+1})] + \eta L \sigma^2
\end{split}
\end{equation}
\begin{equation}
\begin{split}
    &\sum^{M-1}_{m=0}\mathbb{E}\|\nabla f_k(w_{k,p}^{r,m})\|^2_2\\
    &\leq \frac{2}{\eta}\mathbb{E}[f_k(w_{k,p}^{r,0})-f_k(w_{k,p}^{r,M})] + M\eta L \sigma^2
\end{split}
\end{equation}
While from $(r,M)$ to $(r+1,0)$, we have
\begin{equation}
\begin{split}
    &f_k(w_{k,p}^{r+1,0}) - f_k(w_{k,p}^{r,M})\\
    &\leq \langle\nabla f_k(w_{k,p}^{r,M}), w_{k,p}^{r+1,0}-w_{k,p}^{r,M}\rangle + \frac{L}{2}\|w_{k,p}^{r+1,0}-w_{k,p}^{r,M}\|^2_2\\
    &\leq \frac{\eta}{8}\|\nabla f_k(w_{k,p}^{r,M})\|^2_2 + (\frac{2}{\eta} + \frac{L}{2})\|w_{k,p}^{r+1,0}-w_{k,p}^{r,M}\|^2_2\\
    &\leq \frac{\eta}{4}\|\nabla f_k(w^{r,M-1}_{k,p})\|^2_2 + \frac{\eta L^2}{4}\|w^{r,M}_{k,p} - w^{r,M-1}_{k,p}\|^2_2\\
    &\quad + (\frac{2}{\eta} + \frac{L}{2})\|w_{k,p}^{r+1,0}-w_{k,p}^{r,M}\|^2_2\\
    &= \frac{\eta}{4}\|\nabla f_k(w^{r,M-1}_{k,p})\|^2_2 + \frac{\eta^3 L^2}{4}\|\nabla f_k(w^{r,M-1}_{k,p},x^{r,M-1}_{k,p})\|^2_2\\
    &\quad + (\frac{2}{\eta} + \frac{L}{2})\|w_{k,p}^{r+1,0}-w_{k,p}^{r,M}\|^2_2\\
\end{split}
\end{equation}
Therefore, from $(r,0)$ to $(r+1,0)$, we have
\begin{equation}
\begin{split}
    &\sum^{M-1}_{m=0}\mathbb{E}\|\nabla f_k(w_{k,p}^{r,m})\|^2_2\\
    &\leq \frac{2}{\eta}\mathbb{E}[f_k(w_{k,p}^{r,0})-f_k(w_{k,p}^{r+1,0})] + M\eta L \sigma^2\\
    &\quad + \frac{2}{\eta}\mathbb{E}[f_k(w_{k,p}^{r+1,0})-f_k(w_{k,p}^{r,M})]\\
    &\leq \frac{2}{\eta}\mathbb{E}[f_k(w_{k,p}^{r,0})-f_k(w_{k,p}^{r+1,0})] + M\eta L \sigma^2\\
    &\quad + \frac{1}{2}\mathbb{E}\|\nabla f_k(w^{r,M-1}_{k,p})\|^2 + \eta^2L^2 (G^2+\sigma^2)\\
    &\quad + (\frac{4}{\eta^2} + \frac{L}{\eta})\mathbb{E}\|w^{r+1,0}_{k,p}-w^{r,M}_{k,p}\|^2_2
\end{split}
\end{equation}
\begin{equation}
\begin{split}
    &\sum^{M-1}_{m=0}\mathbb{E}\|\nabla f_k(w_{k,p}^{r,m})\|^2_2\\
    &\leq \frac{4}{\eta}\mathbb{E}[f_k(w_{k,p}^{r,0})-f_k(w_{k,p}^{r+1,0})] + 2M\eta L \sigma^2\\
    &\quad + 2\eta^2L^2(G^2+\sigma^2) + (\frac{8}{\eta^2} + \frac{2L}{\eta})\mathbb{E}\|w^{r+1,0}_{k,p}-w^{r,M}_{k,p}\|^2_2
\end{split}
\end{equation}
From $r=0$ to $R-1$,
\begin{equation}
\begin{split}
    &\frac{1}{RM}\sum^{R-1}_{r=0}\sum^{M-1}_{m=0}\mathbb{E}\|\nabla f_k(w_{k,p}^{r,m})\|^2_2\\
    &\leq \frac{4\mathbb{E}[f_k(w^{0,0}_{k,p}) - f_k(w^{R,0}_{k,p})]}{\eta RM} + 2\eta L\sigma^2\\
    &\quad + \frac{2\eta^2L^2(G^2+\sigma^2)}{M} + \frac{1}{RM}(\frac{8}{\eta^2} + \frac{2L}{\eta})\\
    &\quad \cdot\sum^{R-1}_{r=0}\mathbb{E}\|w^{r+1,0}_{k,p}-w^{r,M}_{k,p}\|^2_2\\
    &= \frac{4\mathbb{E}[f_k(w^{0,0}_{k,p}) - f_k(w^{R,0}_{k,p})]}{\eta RM} + 2\eta L\sigma^2\\
    &\quad + \frac{2\eta^2L^2(G^2+\sigma^2)}{M} + \frac{1}{RM}(\frac{8}{\eta^2} + \frac{2L}{\eta})\\
    &\quad \cdot \frac{(1-\lambda)^2 K^2}{(K-1)^2}\sum^{R-1}_{r=0}\mathbb{E}\|w^{r,M}_{k,p}-\overline{w}^{r,M}_{k,p}\|^2_2\\
\end{split}
\end{equation}

\subsection{Proposed Objective}
Here we consider our proposed personalized FL objective function to optimize. For simplicity of notation, let
\begin{equation}
    u^{r,m}_{k}=\frac{1}{\lambda}w_{p,k}^{r,m}-\frac{1-\lambda}{\lambda}\frac{1}{K-1}\sum^{K}_{k^\prime=1,k^\prime\neq k}w_{p,k^\prime}^{r,m}
\end{equation}
Then
\begin{equation}
\begin{split}
    &u^{r,m}_{k} - w^{r,m}_{p,k} = \frac{1-\lambda}{\lambda}(w^{r,m}_{p,k}-\frac{1}{K-1}\sum^{K}_{k^\prime=1,k^\prime\neq k}w_{p,k^\prime}^{r,m})\\
    &= \frac{1-\lambda}{\lambda}\frac{K}{K-1}(w^{r,m}_{p,k}-\overline{w}^{r,m}_{p,k})\\
\end{split}
\end{equation}
Now we bound the gradient of the proposed objective.
\begin{equation}
\begin{split}
    &\frac{1}{K}\sum^{K}_{k=1}\mathbb{E}\|\nabla_{w^{r,m}_{p,k}}\sum^{K}_{k^\prime=1}f_{k^\prime}(u^{r,m}_{k^\prime})\|^2_2\\
    &=\frac{1}{K}\sum^{K}_{k=1}\mathbb{E}\|\frac{1}{\lambda}\nabla f_k(u^{r,m}_{k})-\frac{1-\lambda}{\lambda}\frac{1}{K-1}\nabla f_{k^\prime}(u^{r,m}_{k^\prime})\|^2_2\\
    &\leq \frac{2}{K}\sum^{K}_{k=1} (\frac{1}{\lambda^2}+\frac{(1-\lambda)^2}{\lambda^2(K-1)})\mathbb{E}\|\nabla f_k(u^{r,m}_{k})\|^2_2\\
    &= \frac{2}{K}\sum^{K}_{k=1} (\frac{1}{\lambda^2}+\frac{(1-\lambda)^2}{\lambda^2(K-1)})[\mathbb{E}\|\nabla f_k(w^{r,m}_{k,p})\|^2_2\\
    &\quad + L^2\mathbb{E}\|u^{r,m}_{k}-w^{r,m}_{k,p}\|^2_2]\\
    &= (\frac{1}{\lambda^2}+\frac{(1-\lambda)^2}{\lambda^2(K-1)}) \frac{2}{K}\sum^{K}_{k=1} [\mathbb{E}\|\nabla f_k(w^{r,m}_{k,p})\|^2_2\\
    &\quad +\frac{L^2(1-\lambda)^2K^2}{\lambda^2(K-1)^2}\mathbb{E}\|w^{r,m}_{k,p}-\overline{w}^{r,m}_{k,p}\|^2_2]\\
\end{split}
\end{equation}
\begin{equation}
\begin{split}
    &\frac{1}{KRM}\sum^{R-1}_{r=0}\sum^{M-1}_{m=0}\sum^{K}_{k=1}\mathbb{E}\|\nabla_{w^{r,m}_{p,k}}\sum^{K}_{k^\prime=1}f_{k^\prime}(u^{r,m}_{k^\prime})\|^2_2\\
    &\leq (\frac{1}{\lambda^2}+\frac{(1-\lambda)^2}{\lambda^2(K-1)}) \frac{2}{KRM}\sum^{K}_{k=1}\sum^{R-1}_{r=0}\sum^{M-1}_{m=0}\\
    &[\mathbb{E}\|\nabla f_k(w^{r,m}_{k,p})\|^2_2 + \frac{L^2(1-\lambda)^2K^2}{(K-1)^2}\mathbb{E}\|w^{r,m}_{k,p}-\overline{w}^{r,m}_{k,p}\|^2_2]\\
    &\leq 2(\frac{1}{\lambda^2}+\frac{(1-\lambda)^2}{\lambda^2(K-1)})[\frac{\frac{4}{K}\sum^{K}_{k=1}(f^0_k - f^*_k)}{\eta RM}\\
    &\quad + 2\eta L\sigma^2 + \frac{2\eta^2 L^2(G^2+\sigma^2)}{M}\\
    &\quad + \frac{1}{KRM}(\frac{8}{\eta^2} + \frac{2L}{\eta})\frac{(1-\lambda)^2K^2}{(K-1)^2}\\
    &\quad \cdot \sum^{K}_{k=1}\sum^{R-1}_{r=0}\mathbb{E}\|w^{r,M}_{k,p}-\overline{w}^{r,M}_{k,p}\|^2_2 \\
    &\quad +\frac{1}{KRM}\frac{L^2(1-\lambda)^2K^2}{\lambda^2(K-1)^2}\sum^{K}_{k=1} \sum^{R-1}_{r=0}\sum^{M-1}_{m=0}\mathbb{E}\|w^{r,m}_{k,p}-\overline{w}^{r,m}_{k,p}\|^2_2]\\
\end{split}
\end{equation}
which converges to
\begin{equation}
\begin{split}
    &\mathcal{O}(\frac{1}{\eta RM\lambda^2} + \frac{(1-\lambda)^2}{KRM\eta^2\lambda^2}\sum^{K}_{K=1}\sum^{R-1}_{r=0}\mathbb{E}\|w^{r,M}_{k,p}-\overline{w}^{r,M}_{k,p}\|^2_2\\
    &\quad + \frac{(1-\lambda)^2}{KRM\lambda^4}\sum^{K}_{k=1}\sum^{R-1}_{r=0}\sum^{M-1}_{m=0}\mathbb{E}\|w^{r,m}_{k,p}-\overline{w}^{r,m}_{k,p}\|^2_2)\\
    &=\mathcal{O}(\frac{1}{\eta RM\lambda^2} + \frac{M\sum^{R-1}_{r=0}(1-\lambda)^2}{R\lambda^2}\\
    &\quad + \frac{M^2\eta^2\sum^{R-1}_{r=0}(1-\lambda)^2}{R\lambda^4})
\end{split}
\end{equation}
Suppose $\eta=\mathcal{O}(\frac{1}{\sqrt{RM}})$ and $M=\mathcal{O}(R^{\frac{1}{3}})$, the convergence rate is $\mathcal{O}(\frac{1}{\lambda^4\sqrt{RM}})$ with an error $\mathcal{O}(\frac{M\sum^{R-1}_{r=0}(1-\lambda)^2}{R\lambda^2})$.

\section{Additional Experimental Results}\label{appendix:additional exp}

\begin{figure*}[htb!]
    \centering
    \includegraphics[width=.27\textwidth]{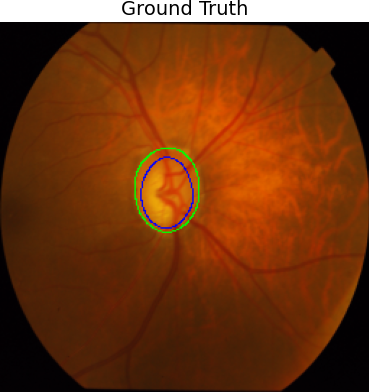}
    \includegraphics[width=.27\textwidth]{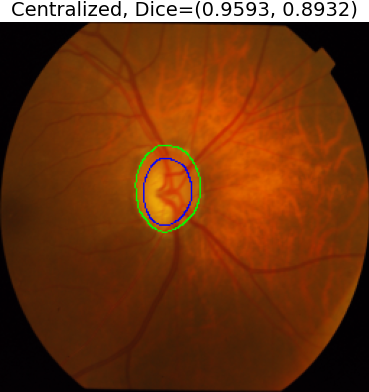}
    \includegraphics[width=.27\textwidth]{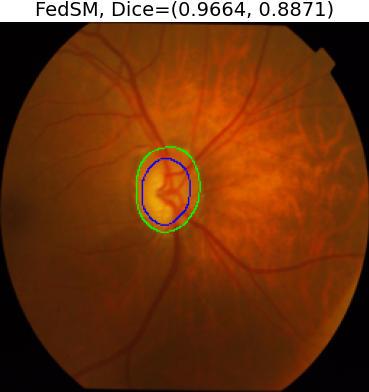}
    \includegraphics[width=.27\textwidth]{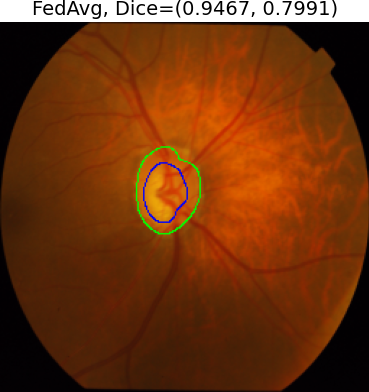}
    \includegraphics[width=.27\textwidth]{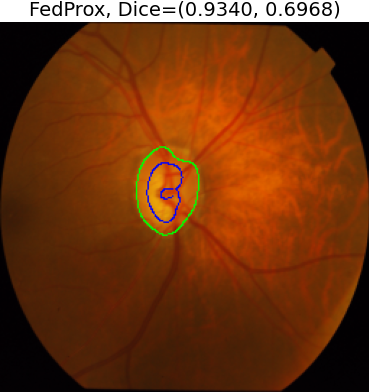}
    \includegraphics[width=.27\textwidth]{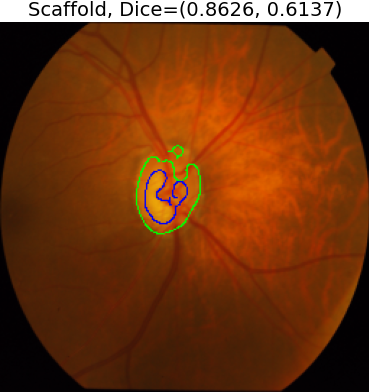}
    \caption{Visual comparison of retinal disc (green) and cup (blue) segmentation. Dice denotes the retinal disc and cup Dice coefficient.}
    \label{fig:visual retinal seg}
\end{figure*}

\begin{figure*}[htb!]
    \centering
    \includegraphics[width=.27\textwidth]{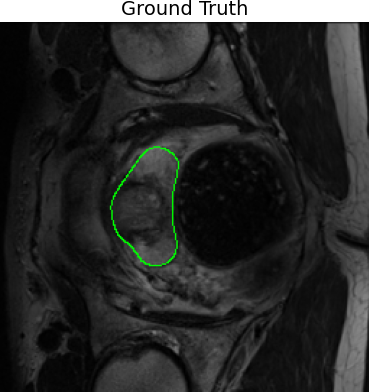}
    \includegraphics[width=.27\textwidth]{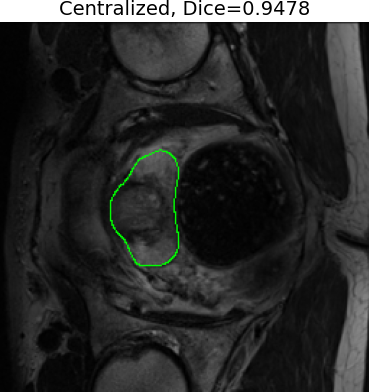}
    \includegraphics[width=.27\textwidth]{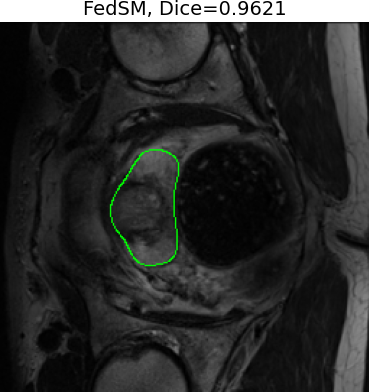}
    \includegraphics[width=.27\textwidth]{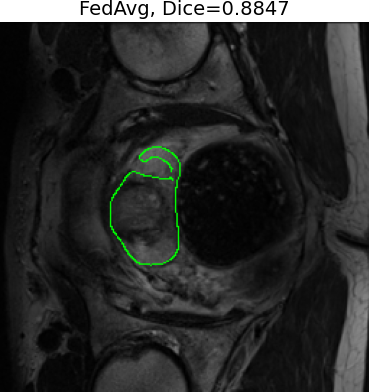}
    \includegraphics[width=.27\textwidth]{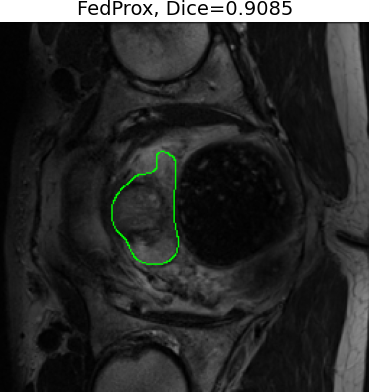}
    \includegraphics[width=.27\textwidth]{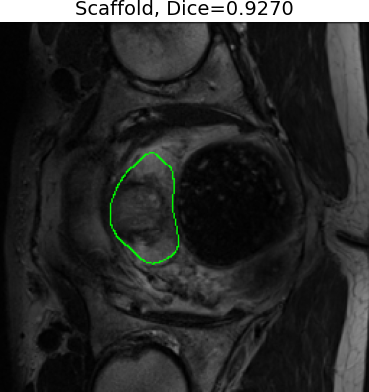}
    \caption{Visual comparison of prostate (green) segmentation. Dice denotes the Dice coefficient.}
    \label{fig:visual prostate seg}
\end{figure*}

\begin{table*}[htb!]
\small
    \centering
    \begin{tabular}{c|cccccc|cc}
        \toprule
        Method & Client 1 & Client 2 & Client 3 & Client 4 & Client 5 & Client 6 & Client Avg Dice & Global Dice \\
        \midrule
        Centralized & 0.9628 & 0.9486 & 0.9489 & 0.9539 & 0.9242 & 0.9565 & 0.9492 & 0.9522 \\
        \midrule
        Client 1 Local & 0.9454 & 0.4357 & 0.8956 & 0.6073 & 0.4464 & 0.8409 & 0.6952 & 0.7129 \\
        Client 2 Local & 0.2936 & 0.9431 & 0.1099 & 0.8371 & 0.2439 & 0.4575 & 0.4809 & 0.5589 \\
        Client 3 Local & 0.9420 & 0.3998 & 0.9468 & 0.6399 & 0.4349 & 0.8256 & 0.6982 & 0.7120 \\
        Client 4 Local & 0.6830 & 0.9400 & 0.4805 & 0.9526 & 0.3088 & 0.7803 & 0.6909 & 0.7796 \\
        Client 5 Local & 0.6102 & 0.2169 & 0.4601 & 0.2518 & 0.9033 & 0.7064 & 0.5248 & 0.5373 \\
        Client 6 Local & 0.8806 & 0.7937 & 0.8354 & 0.8475 & 0.4413 & 0.9547 & 0.7922 & 0.8555 \\
        \midrule
        FedAvg & 0.9554 & 0.9410 & 0.9372 & 0.9535 & 0.8653 & 0.9549 & 0.9346 & 0.9444 \\
        FedProx & 0.9447 & 0.9343 & 0.9229 & 0.9469 & 0.7573 & 0.9480 & 0.9090 & 0.9283 \\
        Scaffold & 0.9207 & 0.9297 & 0.9026 & 0.9474 & 0.6347 & 0.9528 & 0.8813 & 0.9170 \\
        \midrule
        FedSM & \textbf{0.9653} & \textbf{0.9489} & \textbf{0.9545} & \textbf{0.9551} & \textbf{0.9241} & \textbf{0.9560} & \textbf{0.9507} & \textbf{0.9527} \\
        \bottomrule
    \end{tabular}
    \caption{Test Dice coefficient comparison of retinal disc segmentation.}
    \label{tab:retinal disc dice}
\end{table*}

\begin{table*}[htb!]
\small
    \centering
    \begin{tabular}{c|cccccc|cc}
        \toprule
        Method & Client 1 & Client 2 & Client 3 & Client 4 & Client 5 & Client 6 & Client Avg Dice & Global Dice \\
        \midrule
        Centralized & 0.8649 & 0.8033 & 0.8027 & 0.8507 & 0.7778 & 0.8793 & 0.8298 & 0.8507 \\
        \midrule
        Client 1 Local & 0.8216 & 0.2306 & 0.5733 & 0.3793 & 0.2351 & 0.5621 & 0.4670 & 0.4675 \\
        Client 2 Local & 0.1756 & 0.7810 & 0.0673 & 0.7184 & 0.1143 & 0.3637 & 0.3701 & 0.4511 \\
        Client 3 Local & 0.7256 & 0.2807 & 0.8064 & 0.5621 & 0.2939 & 0.7333 & 0.5670 & 0.6068 \\
        Client 4 Local & 0.3385 & 0.7749 & 0.2109 & 0.8491 & 0.1632 & 0.5842 & 0.4868 & 0.6024 \\
        Client 5 Local & 0.4380 & 0.1000 & 0.3305 & 0.1560 & 0.7414 & 0.5380 & 0.3840 & 0.3952 \\
        Client 6 Local & 0.7011 & 0.5360 & 0.6296 & 0.6886 & 0.3073 & 0.8752 & 0.6230 & 0.7198 \\
        \midrule
        FedAvg & 0.8140 & 0.7949 & 0.7963 & 0.8495 & 0.7101 & 0.8795 & 0.8074 & 0.8402 \\
        FedProx & 0.7822 & 0.7702 & 0.7864 & 0.8437 & 0.6132 & 0.8712 & 0.7778 & 0.8216 \\
        Scaffold & 0.7554 & 0.7729 & 0.7405 & 0.8396 & 0.4995 & 0.8732 & 0.7469 & 0.8081 \\
        \midrule
        FedSM & \textbf{0.8610} & \textbf{0.8049} & \textbf{0.8186} & \textbf{0.8530} & \textbf{0.7724} & \textbf{0.8830} & \textbf{0.8322} & \textbf{0.8529} \\
        \bottomrule
    \end{tabular}
    \caption{Test Dice coefficient comparison of retinal cup segmentation.}
    \label{tab:retinal cup dice coefficients}
\end{table*}

\begin{table*}[t]
\small
    \centering
    \begin{tabular}{c|ccccccc|r}
        \toprule
        Unseen Client $k$ & GM & PM1 & PM2 & PM3 & PM4 & PM5 & PM6 & Best $\gamma$, Dice \\
        \midrule
        Client $k=6$ & 1.00 & 0 & 0 & 0 & 0 & 0 & N/A & 1, 0.8906 \\
        Client $k=5$ & 0.69 & 0.18 & 0 & 0 & 0.10 & N/A & 0.03 & 0.9, 0.4304\\
        Client $k=4$ & 0.03 & 0 & 0.97 & 0 & N/A & 0 & 0 & $<$0.95, 0.8870 \\
        Client $k=3$ & 0 & 0 & 0.57 & N/A & 0 & 0 & 0.43 & $<$0.9, 0.8446\\
        Client $k=2$ & 0 & 0 & N/A & 0 & 0.92 & 0.08 & 0 & $<$1, 0.8409 \\
        Client $k=1$ & 0 & N/A & 0 & 1.00 & 0 & 0 & 0 & $<$0.99, 0.8839 \\
        \bottomrule
    \end{tabular}
    \caption{(retinal segmentation, Dice = average of disc and cup Dice coefficients) Model selection frequency from the model selector when FL train with clients $\{1,2,\cdots,6\}/\{k\}$ and test on the \textbf{unseen} client $k\in\{1,2,\cdots,6\}$. From left to right, GM denotes the global model and PM denotes the personalized model $\{1,2,\cdots,6\}/\{k\}$. Here we choose the best $\gamma$.}
    \label{tab:model selector frequency best gamma}
\end{table*}

\begin{table*}[t]
\small
    \centering
    \begin{tabular}{c|cccccc|c}
        \toprule
        Method/Unseen & Client 1 & Client 2 & Client 3 & Client 4 & Client 5 & Client 6 & Avg \\
        \midrule
        Centralized & 0.8842 & 0.8454 & 0.8214 & 0.8866 & 0.4064 & 0.8811 & 0.7875 \\
        FedAvg & 0.8598 & 0.8313 & 0.8224 & 0.8551 & 0.4064 & \textbf{0.8887} & 0.7773 \\
        FedProx & 0.8380 & 0.7856 & 0.8267 & 0.8746 & 0.4171 & 0.8784 & 0.7701 \\
        Scaffold & 0.8085 & 0.7998 & 0.8211 & 0.8568 & 0.4121 & 0.8708 & 0.7615 \\
        \midrule
        FedSM & \textbf{0.8818} & 0.8619 & \textbf{0.8498} & \textbf{0.8901} & 0.4118 & 0.8646 & 0.7933 \\
        FedSM-extra & 0.8747 & \textbf{0.8685} & 0.8467 & 0.8794 & \textbf{0.4265} & 0.8809 & \textbf{0.7963} \\
        \bottomrule
    \end{tabular}
    \caption{(retinal segmentation, Dice = average of disc and cup Dice coefficients) Dice performance when FL train with clients $\{1,2,\cdots,6\}/\{k\}$ and test on the \textbf{unseen} client $k\in\{1,2,\cdots,6\}$.}
    \label{tab:unseen client dice}
\end{table*}

\end{document}